%% file: acl_latex.tex
\pdfoutput=1
\documentclass[11pt]{article}

\usepackage[preprint]{acl}

\usepackage{times}
\usepackage{latexsym}
\usepackage{amsmath,amssymb}
\usepackage{graphicx}
\usepackage[export]{adjustbox}
\usepackage{relsize}
\usepackage{booktabs}
\usepackage{lipsum}
\usepackage{float}
\usepackage{subcaption}
\usepackage{comment}
\usepackage{cleveref}
\usepackage{array}
\usepackage{bbm}
\usepackage{multirow,multicol}
\usepackage[round-mode=places,round-precision=4]{siunitx}
\usepackage{ragged2e}

\DeclareMathOperator*{\argmax}{argmax}

\usepackage[T1]{fontenc}
\usepackage[utf8]{inputenc}

\usepackage{microtype}

\usepackage{inconsolata}

\usepackage{graphicx}

\title{\textcolor{black}{Your Model is Overconfident, and Other Lies We Tell Ourselves}}

\author{
{\bf Timothee Mickus\textsuperscript{1} \qquad Aman Sinha\textsuperscript{2} \qquad Raúl Vázquez\textsuperscript{1}}
\\
 \textsuperscript{1}University of Helsinki \qquad \textsuperscript{2}Université de Lorraine
\\
\texttt{firstname.lastname@\{}\textsuperscript{1}\texttt{helsinki.fi,}\textsuperscript{2}\texttt{univ-lorraine.fr\}}
}

\begin{document}
\maketitle
\begin{abstract}
The difficulty intrinsic to a given example, rooted in its inherent ambiguity, is a key yet often overlooked factor in evaluating neural NLP models.
We investigate the interplay and divergence among various metrics for assessing intrinsic difficulty, including annotator dissensus, training dynamics, and model confidence. Through a comprehensive analysis using 29 models on three datasets, we reveal that while correlations exist among these metrics, their relationships are neither linear nor monotonic. By disentangling these dimensions of uncertainty, we aim to refine our understanding of data complexity and its implications for evaluating and improving NLP models.
\end{abstract}

\section{Introduction}

A central, but often overlooked, concept in natural language processing is the consensus of annotators when labeling a specific datapoint. 
Annotators often express different perspectives, and prior literature provides strong evidence that this dissensus is a legitimate characteristic of language data, rather than a consequence of noisy annotation processes \citep{plank-etal-2014-linguistically,plank-2022-problem,Uma2021}.

\textit{Annotator dissensus} is often linked to \textit{data complexity}: if humans do not agree as to whether a pair of sentences contradict each other, then we expect this pair to be hard to label with a neural network.
Data complexity can be quantified through training dynamics (e.g., how early a training example is learned), model confidence, or performance variability across models \citep{pmlr-v70-guo17a,swayamdipta-etal-2020-dataset, hendrycks2018benchmarking}. This relationship shapes the design of evaluation benchmarks, which increasingly incorporate multiple annotations per datapoint \citep[e.g.,][]{bowman-etal-2015-large,nie-etal-2020-learn}. 
% A specifically relevant example is the ChaosNLI dataset of \citet{nie-etal-2020-learn}, which collects 100 annotation for a few thousand pairs of premises and hypotheses drawn from earlier NLI validation set.
As such, matching human uncertainty with model uncertainty is an explicit desideratum laid out in numerous NLP applications.

Common drivers of data uncertainty include annotation noise, semantic ambiguity, and overlapping class boundaries \citep{hu2023uncertainty}. These factors not only lower inter-annotator agreement but also highlight the presence of linguistic phenomena such as semantic indeterminacy --- which lead to label ambiguity, as extensively documented across NLP tasks \citep{bowman-etal-2015-large}. % Furthermore, it is worth stressing what causes are generally proposed as the underlying driving force towards greater data uncertainty: For instance, \citet{hu2023uncertainty} include noise in the collected annotations, ambiguity of the data presented, or overlap between different valid classes among the root causes for data complexity. It is reasonable to expect that annotator dissensus is also aggravated when such factors are more salient: noisy annotation procedures should lower inter-annotator agreement; and phenomena such semantic indeterminacy leading to label ambiguity are well documented across many NLP task \citep{bowman-etal-2015-large}.
Hence, prior work has expected annotator dissensus to be a reasonable proxy for data complexity \citep[e.g.,][]{hachey-etal-2005-investigating,Lalor2018-ft}.

Our findings %challenge this assumption, 
reveal non-linear and conflicting relationships between annotator dissensus and model-derived complexity metrics.
Through experiments using 29 models on the ChaosNLI and DynaSent datasets \citep{nie-etal-2020-learn,potts-etal-2021-dynasent}, we observe that various metric indicators of data complexity often correlate with one another  --- and yet that the relationship they hold with human-based assessments of linguistic dissensus is far from linear or monotonic.
Moreover, indicators of data complexity derived from model behavior tend to conflict depending on whether they account for the correctness of the models' predictions --- for instance, assessments derived from conformal prediction methods do not align with the training dynamics approach of \citet{swayamdipta-etal-2020-dataset}, and both fail to adequately capture the true human label variation collected by \citet{nie-etal-2020-learn}.
% In short, we can identify a subset of datapoints where models overwhelmingly agree on a label that does not match the consensus among annotators.

\section{Background}
\input{background_new}
\section{Indicators of data complexity}
\label{sec:indicators}

%In the present work, w
We formalize several means of assessing how difficult it is to assign one of $k$ possible labels $\left \{y_1, \dots, y_k \right \} = Y$ to a specific instance $x$. 

\subsection{Human-based indicators}

\paragraph{Empirical population dissensus.}
The simplest way to quantify disagreement on a specific datapoint is to ask multiple annotators $a_1, \dots, a_n$, and compute how unpopular the majority opinion is. 
As such, if annotator $a_j$ would assign the label $y_{a_j}$ to the observation $x$, we can define the probability  $\Pr_\mathbb{H}(y_i | x)$ on label $y_i$ as the proportion of annotators agreeing on the label $y_i$, or formally
$$\Pr{}_\mathbb{H}^{}(y_i | x) = \frac{1}{n}\sum \limits_{j=1}^n \mathbbm{1}\left\{y_{a_j} = y_i\right\}$$
and denote the dissensus among annotators as:
\begin{equation}
    \mathbb{H}_\mathrm{dis} = 1 - \max\limits_{y_i \in Y} \Pr{}_\mathbb{H}^{}(y_i | x),
    \label{eq:human dis}
\end{equation}
% where $|\cdot|$ denotes the cardinality of its argument. 
$\mathbb{H}_\mathrm{dis}$ is hence inversely related to the popularity of the most common label. If all annotators agree, there is a strong consensus, implying that $\mathbb{H}_\mathrm{dis}=0$ because $\max\limits_i \Pr{}_\mathbb{H}^{}(y_i | x)=1$.
% The simplest way to quantify annotators disagreement for a specific datapoint is to compute the
\paragraph{Empirical population entropy.}
The empirical dissensus $\mathbb{H}_\mathrm{dis}$ has the drawback of not factoring in minority opinions: there is a distinction to be made between having the opinions split among a handful of well-supported alternatives versus a total lack of consensus and annotators maximally split across all possible alternatives.
To account for such differences, we consider the empirical entropy of opinions \citep{nie-etal-2020-learn}, or 
\begin{equation}
    \mathbb{H}_\mathrm{ent} =  - \sum\limits_{y_i \in Y} \Pr{}_\mathbb{H}^{}(y_i | x) \log \Pr{}_\mathbb{H}^{}(y_i | x) 
    \label{eq:human ent} 
\end{equation}
%where $p_\mathrm{ann}(y_i | x)$ is the empirical probability of any  annotator associating point $x$ with label $y_i$. 
Entropy measures uncertainty or diversity in the label distribution, better accounting for both dominant and minority labels. As before, $\mathbb{H}_\mathrm{ent} = 0$ when all annotators agree on one label. Contrastingly, $\mathbb{H}_\mathrm{dis}$ is maximal when $\Pr_\mathbb{H}^{} \sim \mathrm{Unif}$, i.e., when annotators are evenly split across all labels.
\subsection{Reference-free model-based indicators}

\paragraph{Model pool dissensus and model pool entropy.}
Given a set of models parametrized by $\theta_1, \dots, \theta_m$, we can easily extend the concepts of dissensus ($\mathbb{H}_\mathrm{dis}$) and entropy ($\mathbb{H}_\mathrm{ent}$) to models' predictions, instead of relying on human annotators. 
To do this, we evaluate the predictions of a model $\theta_j$ by selecting the label  $\argmax_{y_i \in Y}~p(y_i | x, \theta_j)$. 
Next, we define the probability $\Pr_\mathbb{M}(y_i | x)$ of this datapoint being labeled as $y_i$, by tallying the number of models that predict $y_i$ as the most likely label:
\[\resizebox{\columnwidth}{!}{$\Pr{}_\mathbb{M}^{}(y_i | x) = \frac{1}{m}\sum \limits_{j=1}^m \mathbbm{1}\Bigl\{ y_i = \argmax\limits_{y_k \in Y}~p(y_k | x, \theta_j) \Bigr\}$}\]
Using this %probability 
distribution, we can analogously define both metrics for the models' predictions:
  \begin{align}
    \mathbb{M}_\mathrm{dis} &= 1 - \max\limits_{y_i \in Y} \Pr{}_\mathbb{M}^{}(y_i | x)
    \label{eq:model-ref dis} \\
      \mathbb{M}_\mathrm{ent} &=  - \sum\limits_{y_i \in Y} \Pr{}_\mathbb{M}^{}(y_i | x) \log \Pr{}_\mathbb{M}^{}(y_i | x)
   \label{eq:model-ref pop ent}
  \end{align}
%this time using the observed probability of a given model prediction $p_\mathrm{mod}(y_i | x) = \frac{|M_{ij}|}{m}$.

\paragraph{Averaged model entropy.}
Entropy has also been used to assess the confidence of a model in its own prediction \citep[e.g.,][]{malinin2021uncertainty,schroder-etal-2022-revisiting,baumler-etal-2023-examples}.
A reasonable line of thought is that lower confidence scores reflect data complexity.
To evaluate the difficulty of labeling $x$, we can average the label distribution entropy across multiple %pool of 
models:
\begin{align}
    \mathbb{M}_\mathrm{avg\ ent} = -\frac{1}{m} \sum\limits_{j=1}^m \sum\limits_{y_i \in Y}  & p(y_i | x, \theta_j) \nonumber \\ & \times \log p(y_i | x, \theta_j)
    \label{eq:model-ref avg ent}
\end{align}

\paragraph{Conformal prediction set size.}
A more elaborate statistical estimator than entropy consists in quantifying the ambiguity necessary for a probabilistic classifier to meet a certain statistical guarantee; an approach known as conformal prediction (CP, \citealp{vovk-etal-2005-algorithmic,angelopoulos2022gentle}).
In practice, we can also use a probabilistic classifier parametrized with $\theta$ to derive a set of possible labels $\mathcal{C}_\theta(x) \subseteq Y$  for every input $x$ such that the true label $y^\ast$ is likely to be in $\mathcal{C}_\theta(x)$, with a budget tolerance for failure $1-\alpha$. % --- or 
Formally, we want to construct a set-valued function $\mathcal{C}_\theta$ such that %for all $x$, 
$$\forall x \qquad \Pr\left(y^\ast \in \mathcal{C}_\theta(x)\right) \geq 1 - \alpha$$
% where $\Pr$ is understood in a frequentist sense.
We can then capture the ambiguity inherent to a prediction by considering the size of the prediction set, $|\mathcal{C}_\theta(x)|$: a larger CP set size ought to reflect a greater uncertainty as to what the true label is.
To convert a probabilistic classifier $p(Y|X, \theta)$ to such a set-valued classifier, we rely on a least-ambiguous set-valued classifier method \citep{doi:10.1080/01621459.2017.1395341}.
This consists in identifying the value $t_\theta$ such that, for all calibration datapoints $x'$ with their label $y'$ in a held-out calibration dataset $\mathcal{D}_\mathrm{cal}$:
\begin{align*}
  \hat{q} &= \frac{|\mathcal{D}_\mathrm{cal}| + 1}{|\mathcal{D}_\mathrm{cal}|} (1 - \alpha) \\
  t_\theta &= \sup \left\{t \middle| \Pr\left(p\left(y' | x', \theta\right) \geq t\right) \geq \hat{q} \right\} 
\end{align*}
Using $t_\theta$, we can construct the set 
$$\mathcal{C}_\theta(x) = \left\{y ~ \middle| ~ p\left(y | x, \theta\right) \geq t_\theta \right\}$$ 
which provides the expected statistical guarantee.
Here, we convert CP sets into uncertainty indicators by considering their average size across models:
\begin{equation}
    \mathbb{M}_\mathrm{CP} = \frac{1}{m}\sum\limits_{i=1}^m \left|\mathcal{C}_{\theta_i}(x)\right| \label{eq:model-ref cp}
\end{equation}
Here, we experiment with three variants, based on different risk tolerances with $\alpha=0.05$, $\alpha=0.1$ or $\alpha=0.2$. %; we denote these variants as $\mathbb{M}_{\mathrm{CP\ }\alpha=0.05}$, $\mathbb{M}_{\mathrm{CP\ }\alpha=0.1}$ and  $\mathbb{M}_{\mathrm{CP\ }\alpha=0.2}$.
While conformal prediction algorithms require labeled calibration sets $\mathcal{D}_\mathrm{cal}$, their predictions are made without label information. Hence we consider CP set size indicators to be reference-free, as they can estimate uncertainty for unlabeled datapoints.
We use as $\mathcal{D}_\mathrm{cal}$ all other datapoints in the test set (i.e., a leave-one-out process).
% Furthermore, our experiments below also demonstrate they are more in line with reference-free indicators than reference-dependent indicators.

\subsection{Reference-dependent model-based indicators}

\paragraph{Model pool failure rate.} Since it is in principle possible for models to broadly agree on a label that human annotators would not have picked, one value worth considering is the proportion of models that fail to produce the reference label $y^\ast$ we would expect given our annotations. Defining
\begin{equation}
    \mathbb{M}_\mathrm{fail}^{\mathrm{ref}} = \frac{1}{m} \sum\limits_{i=1}^m \mathbbm{1}\Bigl\{\argmax _{y_j\in Y} p(y_j | x, \theta_i) \neq y^\ast\Bigr\}
    \label{eq:model+ref fail}
\end{equation}
highlights the disconnect between model predictions and human annotations. A low value for $\mathbb{M}_\mathrm{fail}^{\mathrm{ref}}$ implies a strong alignment between the model pool and the human-provided reference label; a high value suggests that many models fail to predict 
$y^\ast$. % However, a high model-pool failure rate  does not necessarily reflect inherent ambiguity in the input point $x$, but rather limitations or biases in the models themselves --- especially when human annotators unanimously agree on $y^\ast$.

\paragraph{Early computation termination.}
\citet{baldock2021deep} propose to estimate the difficulty of an example through the computational cost of a correct prediction.
%Their approach consists in verifying at which layer activations start to form neighborhoods that are homogeneous with respect to their labels:
% In practice, \citeauthor{baldock2021deep} 
They first compute the hidden representations $\mathbf{h}_i^1, \dots \mathbf{h}_i^l$ for a specific input $\mathbf{x}_i$ and then assess which of these  representations lie in label-specific subspaces using kNN classifiers, 
% Their hypothesis is 
since that datapoints that are easier to process ought to be mapped onto unambiguous subspaces earlier. % than complex ones. 

% The approach of \citeauthor{baldock2021deep} has the downside that it
This approach assumes there is a meaningful distance metric between the different representations --- an assumption that is not easy to meet with sequence-level classification tasks, where inputs can have different matrix shapes.
We can however leverage the fact that Transformer layers can be viewed as functions mapping from and unto the same space \citep{elhage-etal-2021-circuits-thread}: Earlier work has suggested to interpret hidden representations for a specific layer by directly projecting them onto the label-space, skipping over all subsequent layers \citep{nostalgebraist-2020-logit-lens,geva-etal-2022-transformer}.
%In practice, we start with the remark that most modern neural networks are \emph{deep} --- i.e., they correspond to a stack of layers with homologous definitions but distinct parametrization.
%A consequence of that is that we can probe the behavior of a deep neural net by simply ignoring all subsequent layers, and directly projecting these representations onto the label space.
We therefore replace \citeauthor{baldock2021deep}'s kNN classifiers with the learned classifier head. 
More formally, if a model parametrized with $\theta_i$ is made of $l_i$ layers of the form $f_{\theta_i,j}(\mathbf{X}) = \phi(\mathbf{X}, \theta_{i,j})$ and a projection head $f_{\theta_i,\mathrm{proj}}(\mathbf{X}) = \argmax \psi(\mathbf{X}, \theta_{i,l_i+1})$,  let us denote all early predictions from layer $j$ onward as 
\[\resizebox{\columnwidth}{!}{$\hat{Y}_{ij} = \Bigl\{f_{\theta_i,\mathrm{proj}}\circ f_{\theta_i,k}\ \circ \dots \circ f_{\theta_i,1} \left(\mathbf{X}\right) ~\Bigm|~ j \leq k \leq l_i \Bigr\}$}\]
%using which we can consider the first layer from which all predictions onward are correct, according to a reference label $y^\ast$:
which allows us to retrieve the first layer $k$ such that all predicts from layer $k$ to layer $l$ are  correct, according to a reference label $y^\ast$:
\begin{align}
    \mathrm{s}_{1^\mathrm{st}\ \mathrm{layer}}^{\mathrm{ref}}(\theta_i) &= \begin{cases}
        1 \hfill \mathrm{if}~p(y | x, \theta_i) \neq y^\ast \\
        \frac{\min\limits_j \left\{j ~\middle|~  \hat{Y}_{ij} = \{y^\ast \}\right\}}{l+1} \quad \mathrm{otherwise}
    \end{cases} 
    \nonumber \\
    \mathbb{M}_{1^\mathrm{st}\ \mathrm{layer}}^{\mathrm{ref}} &= \frac{1}{m} \sum\limits_{i=1}^m \mathrm{s}_{1^\mathrm{st}\ \mathrm{layer}}^{\mathrm{ref}}(\theta_i)
    \label{eq:model+ref 1st layer}
\end{align}

We average across our pool of models so that the indicator is not too sensitive to one specific model's idiosyncratic behavior.
This also leads us to normalizing according to the number of layers so that we maintain consistent ranges across models with different layer counts.
We also make the practical choice of setting examples that models do not label correctly to the higher end of the scale.

\paragraph{Early training termination.}
One can also consider that easier items require less training \citep{swayamdipta-etal-2020-dataset}.
If for a given model $\theta_i$ we have access to different checkpoints across training $\theta_i^1, \dots, \theta_i^p$, we can simply assess when the model starts making reliable predictions. 
Consider the set of predictions from all future checkpoints:
\[\resizebox{\columnwidth}{!}{$F_{ij} = \left\{\argmax\limits_y p(y | x, \theta_i^j), ~\dots,~ \argmax\limits_y p(y | x, \theta_i^p)\right\}$}\]
which we use to define:
\begin{align}
    \mathrm{s}_{1^\mathrm{st}\ \mathrm{ckpt}}^{\mathrm{ref}}(\theta_i^p) &= \begin{cases}
        1 \hfill \mathrm{if}~p(y | x, \theta_i^p) \neq y^\ast \\
        \frac{\min\limits_j \left\{j ~\middle|~ F_{ij}  =  \left\{y^\ast \right\} \right\}}{p+1} \quad \mathrm{otherwise}
    \end{cases} \nonumber \\
    \mathbb{M}_{1^\mathrm{st}\ \mathrm{ckpt}}^{\mathrm{ref}} &= \frac{1}{m} \sum\limits_{i=1}^m \mathrm{s}_{1^\mathrm{st}\ \mathrm{ckpt}}^{\mathrm{ref}}(\theta_i^p)
    \label{eq:model+ref 1st ckt}
\end{align}
Here again, we normalize according to the number of checkpoints, average across all models, and manually penalize models that do not ultimately learn to produce the target reference.

\paragraph{Failure rate through training.}
We can also assume that easier items are likely to be attributed the expected reference at any stage of training, whereas more complex observations will only be labeled properly during the later stages. 
We can therefore quantify the proportion of checkpoints where the model failed to produce the expected label $y^\ast$:
\begin{equation}
    % \mathrm{s}_\mathrm{avg\ ckpt}^{\mathrm{ref}}&(\theta_i^j) = \begin{cases}
    %     1 \qquad \mathrm{if}~p(y | x, \theta_i^j) \neq y^\ast \\
    %     0 \qquad \mathrm{otherwise}
    % \end{cases}\nonumber \\
    \mathbb{M}_\mathrm{avg\ ckpt}^{\mathrm{ref}} = \frac{1}{mp} \sum\limits_{i=1}^m \sum\limits_{j=1}^p  \mathbbm{1}\left\{p(y | x, \theta_i^j) \neq y^\ast\right\} %\mathrm{s}_\mathrm{avg\ ckpt}^{\mathrm{ref}}(\theta_i^j) 
    \label{eq:model+ref avg ckt}
\end{equation}
Again, we average across a pool of models to mitigate idiosyncrasies.

\paragraph{Probability mass through training.}
One problem with the approach in \cref{eq:model+ref avg ckt} is that it does not distinguish between cases where the classifier correctly predicts $y^\ast$ and assigns no weight to any other options from cases where the probability assigned to $y^\ast$ is only within a small margin from that of an incorrect class.
% \footnote{
%     In practice, most of our reference-dependent model-based indicators (esp. \cref{eq:model+ref fail}) could be likewise softened by considering the model estimate probability mass. This however begs the question of model calibration \citep{guo_calibration_2017}. %i.e., there are limited guarantees that the probability estimates yielded by a model reflect a frequentist or Bayesian conception of a probability distribution.
% }
\citet{swayamdipta-etal-2020-dataset} propose to consider the probability mass assigned by the classifier across training,
% \footnote{
%     This indicator corresponds to what \citeauthor{swayamdipta-etal-2020-dataset} call ``confidence.''
% } 
or formally:
\begin{equation}
    \mathbb{M}_{\mathrm{avg\ ckpt\ }p}^{\mathrm{ref}} =  1 - \frac{1}{mp} \sum\limits_{i=1}^m \sum\limits_{j=1}^p p(y^\ast | x, \theta_i^j)
    \label{eq:model+ref avg ckt p}
\end{equation}
Eq.~(\ref{eq:model+ref avg ckt p}) is minimized when the gold label $y^\ast$ is assigned a probability of 1 throughout training.

\section{ChaosNLI}
\label{sec:chaosNLI}
\subsection{Experimental setup}

% In this work, we study classifiers trained on SNLI \citep{bowman-etal-2015-large} or MNLI \citep{williams-etal-2018-broad} and evaluated on ChaosNLI \citep{nie-etal-2020-learn}.
We first study classifiers trained on SNLI \citep{bowman-etal-2015-large} or MNLI \citep{williams-etal-2018-broad} and evaluated on ChaosNLI \citep{nie-etal-2020-learn}.
We might expect the family of models we consider to define our indicators to weigh on results. % in \cref{eq:model-ref dis,eq:model-ref pop ent,eq:model-ref avg ent,eq:model-ref cp,eq:model+ref fail,eq:model+ref 1st layer,eq:model+ref 1st ckt,eq:model+ref avg ckt,eq:model+ref avg ckt p}.
In particular, the homogeneity of the pool of models considered --- in terms of pretraining and fine-tuning data, algorithmic and architectural designs, or parameter counts --- is a factor of interest.
% As such, we consider two groups of PLM-based classifiers.

\paragraph{Heterogeneous training, similar parameter counts (1B group).} 
One may expect that data complexity indicators should be established by considering a large swath of models trained in conditions as varied as possible --- i.e., using different training data and algorithms. 
To this end, we consider 5 different LLMs in the 1B parameter range; OLMo \citep{groeneveld-etal-2024-olmo}, Pythia \citep{biderman2023pythia}, Llama 3.2 \citep{grattafiori2024llama3herdmodels}, Falcon \cite{almazrouei2023falcon}, and BLOOM \citep{le2023bloom}.
So as to further maximize the difference across the different models we consider, we partition the NLI training set (either SNLI or MNLI) into five equally sized subsets $s_1, \dots, s_5$ and train one model for each pair of LLM and NLI subset, or 25 classifiers on SNLI and MNLI each.
% We then measure performances on the SNLI portion of ChaosNLI.
%TODO hparam details

\paragraph{Homogeneous training data, different parameter counts (<1B group).} 
Conversely, we might expect that the model pool should be established with a fixed training data --- on the one hand, this corresponds to an assumption frequently made when measuring aleatoric uncertainty in the Bayesian literature; %TODO refs
on the other hand, we might expect that difficulty should be intimately linked to the data a model has been exposed to. %TODO refs
To that tend, we consider a family of smaller BERT-type models \citep{turc2019} so as to verify how the indicators in  behave with respect to a family of different models trained homogeneously on the same data and under the same conditions, varying in terms of architecture designs and parameter counts.
% \footnote{
%     For our study, we rank these models with respect to     \emph{complexity} and denote them via $m_i$ where $i$ is their rank in Table \ref{tab:complexmodels} for simplicity. $i$ ranges from 1 to 29, where 1 to 24 are BERT miniature models and 25 or later are LLMs. $\theta_1$ is the model with least number of parameters in the collection and $\theta_{29}$ is the model with highest number of parameters.
% }
We fine-tune all of \citeauthor{turc2019}'s BERT models on %TODO a subset of
each of the NLI training sets in their entirety. %; we measure performances on the relevant section of ChaosNLI. 

% \paragraph{Dataset.} We train all the above mentioned configurations for SNLI dataset and further use ChaosNLI dataset to access multiple annotations of part of($\sim$1500 samples).

\subsection{Results}

\paragraph{Human-based and model-based indicators do not agree with each other.}
A straightforward first approach consists in computing how the different indicators correlate with one another --- in particular, we start by focusing on comparing human-based indicators %(\cref{eq:human dis,eq:human ent}) 
to model-based indicators. % (\cref{eq:model-ref dis,eq:model-ref pop ent,eq:model-ref avg ent,eq:model-ref cp,eq:model+ref fail,eq:model+ref 1st layer,eq:model+ref 1st ckt,eq:model+ref avg ckt,eq:model+ref avg ckt p}).

\input{Tables/snli-results/snli-spearman}

\input{Tables/mnli-results/mnli-spearman}

The corresponding Spearman correlation values are shown in \Cref{tab:snli:corrtable,tab:mnli:corrtable}. %, with \Cref{tab:corrtable-bert} displaying measurements for the BERT-based (<1B) models and \Cref{tab:corrtable-1b} for the heterogeneous LLM-based (1B) models.
% We can make two remarks. 
% First, and p
% Perhaps unsurprisingly, r
Reference-free indicators defined without factoring in the majority label among human annotators (\cref{eq:model-ref dis,eq:model-ref pop ent,eq:model-ref avg ent,eq:model-ref cp}) almost systematically yield lower correlations than reference-dependent indicators (\cref{eq:model+ref fail,eq:model+ref 1st layer,eq:model+ref 1st ckt,eq:model+ref avg ckt,eq:model+ref avg ckt p}).

%Second, 
Yet, while we observe positive and significant trends throughout, the correlation itself is somewhat low. 
For a sense of scale, if we are to focus on SNLI for which we observe the highest correlations,  two human-based indicators or two reference-dependent indicators, tend to yield correlation scores of $\rho \geq 0.9$.
When comparing two reference-free indicators, we can observe two sub-groups: namely, $\mathbb{M}_\mathrm{avg\ ent}$ and the CP set size indicators yield correlations of $\rho \geq 0.9$),\footnote{
    Except $\mathbb{M}_{\mathrm{CP\ }\alpha=0.05}$ and $\mathbb{M}_{\mathrm{CP\ }\alpha=0.2}$, where $\rho \approx 0.80$.
} whereas $\mathbb{M}_\mathrm{dis}$ and $\mathbb{M}_\mathrm{ent}$ yield a correlation of $\rho \approx 0.95$, and comparisons across these two sub-groups are in the range $0.64 < \rho < 0.88$.
The observation also holds on MNLI: We observe a correlation of $\rho \approx 0.90$ for $\mathbb{H}_\mathrm{dis}$ and $\mathbb{H}_\mathrm{ent}$, correlations systematically greater than $\rho \geq 0.9$ between any two reference-dependent indicators, and correlations between $0.46 \leq \rho \leq 0.96$ for reference-free indicators (with again $\mathbb{M}_\mathrm{dis}$ and $\mathbb{M}_\mathrm{ent}$ forming a subgroup distinct from $\mathbb{M}_\mathrm{CP}$ and $\mathbb{M}_\mathrm{avg\ ent}$).
In sum, all three groups of indicators %described in \S\ref{sec:indicators} 
portray different pictures, echoing findings from prior works (esp. \citealp{pavlick-kwiatkowski-2019-inherent}): 
The difficulty associated to the samples is \emph{not} the same for the humans and models, regardless of the pool considered.\footnote{
    We can stress this relationship is non-linear, see \S\ref{adx:sup-res:residual analysis}.
}

The behavior of model-based indicators also appears contingent on the exact setup.
For instance, observations derived from our 1B model pool on MNLI would suggest $\mathbb{M}_{\mathrm{CP\ }\alpha=0.1}$ to be quite in line with human label variation assessments --- whereas the corresponding coefficient in the <1B pool on MNLI is about 0. 
In the same vein, the choice of $\alpha$ for CP has different effects on SNLI and MNLI insofar the 1B pool is concerned: Whereas  $\mathbb{M}_{\mathrm{CP\ }\alpha=0.2}$ yields higher results than  $\mathbb{M}_{\mathrm{CP\ }\alpha=0.05}$ on MNLI, the opposite is true for SNLI classifiers.

\begin{figure}
    \centering
    \includegraphics[max width=0.9\columnwidth, trim={0 0.35cm 0 0.225cm}, clip]{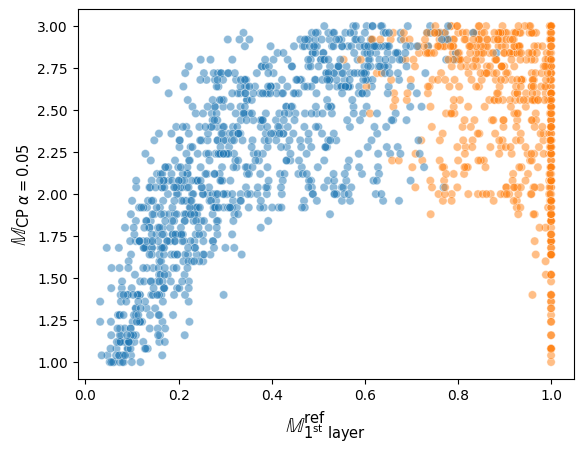}
    \caption{Example of joint distribution between a reference-free and a reference-dependent indicator (SNLI 1B pool, $\mathbb{M}_{\mathrm{CP,\ }\alpha=0.05}$ vs. $\mathbb{M}_{1^\mathrm{st}\mathrm{\ layer}}^\mathrm{ref}$). Datapoints in orange are misclassified by $50\%$ of the pool, blue datapoints aren't. See also \cref{tab:snli:corrtable-models-partitioned,tab:mnli:corrtable-models-partitioned} (\S\ref{adx:sup-res:breakdown along success}).}
    \label{fig:example-model-v-model}
\end{figure}

\paragraph{Reference-free indicators conflate model success and model failure.}
We can also remark that reference-free and reference-dependent indicators do not agree either.
This is evident, for instance, by looking at \Cref{fig:example-model-v-model}, which exemplifies one such comparison.
We can see that the joint distribution of the indicators forms an inverted U-shape distribution, i.e., the reference-free indicator rates as equally good items that the reference-dependent does discriminates.
Generally speaking, reference-free indicators tend to assign similar scores to datapoints rated as either maximal or minimal by reference-dependent indicators:
In fact, if we partition datapoints according to whether a majority of the models fail to predict the annotator majority label (corresponding to the orange and blue hues in \Cref{fig:example-model-v-model}), we can observe systematic \emph{anti}-correlations when the models do tend to fail.\footnote{
    See in \Cref{tab:snli:corrtable-models-partitioned,tab:mnli:corrtable-models-partitioned} in \S\ref{adx:sup-res:breakdown along success} for detailed results.
}
One major factor at play here is that models fail more often on samples with a high human dissensus.
This can be shown with Mann-Whitney U tests.
On SNLI, for the 1B models, we observe a $p$-value of $p < 10^{-27}$ and a common language effect size $f=66.7\%$; as for the <1B models, we have $p < 10^{-42}$, $f=72.2\%$.
On MNLI, the 1B model pool yields $p < 10^{-14}$ and $f=61.3\%$, whereas the <1B pool yields $p<10^{-7}$ and $f=58.1\%$.\footnote{
    Conversely, we can still identify a small subset of data\-points with low human dissensus but high model failure rates.
}

{\color{black}
\section{DynaSent}
\label{sec:dynasent}
\subsection{Experimental setup}

% To be written properly: the point with DynaSent is to test (i) whether we replicate our findings on another task than NLI, and (ii) whether we can throw a bone at R3 and figure out if finetuning the classifiers to match the distribution of labels (rather than only the majority opinion) actually changes the picture. 

An overlooked aspect of our discussion so far is whether the discrepancy between indicators highlighted in \S\ref{sec:chaosNLI} can be mitigated. 
Since we trained our classifiers to match the consensus among annotators, any probability mass not assigned to the majority label is penalized.
As a result, classifiers default to assigning most of their probability mass to a single label. That is, they may be poorly calibrated in the sense of \citet{pmlr-v70-guo17a}, and the probabilities that they assign might not reflect the probability of the model being correct.
As \citet{baan-etal-2022-stop} discuss, measuring calibration for ambiguous labels is not possible in practice as metrics such as ECE explicitly assume the existence of a single valid label. \citeauthor{baan-etal-2022-stop} instead advocate to measure how well the model's distribution aligns with human label variation.\footnote{
    This is similar to what we present in \Cref{tab:snli:corrtable,tab:mnli:corrtable}.
} % of calibration in the context of human label variation.
We thus expect that directly optimizing the classifiers to match the empirical human label variation will remedy this issue. We refer to this approach as training with `soft' continuous labels, instead of `hard' categorical labels. 

% DynaSent \citep{potts-etal-2021-dynasent} is a ternary (positive/negative/neutral) sentiment analysis dataset. In total, it contains a total of 121,634 sentences, each validated by five crowdworkers. However, due to occasional undecisive human annotations, we created two subset of the datasets: soft-subset, and hard-subset. In [add subset name] we remove all cases where any one of the 5 annotators marked the datapoint as "mixed", this leads to reduction of the training set by 30\% (See Appendix \cref{app:dyna-stats}). And, for [add subset name], we do ... 

For these experiments we use DynaSent \citep{potts-etal-2021-dynasent}, which also allows us to assess whether out results generalize beyond NLI. 
DynaSent is a sentiment analysis dataset built incrementally through several rounds of adversarial data collection,  featuring four sentiment classes: positive, negative, neutral or mixed.  
Each datapoint is annotated by five crowd-workers, providing a first-order approximation of human label variation.
We focus on the round 1 data to train classifiers using either hard or soft labels as targets.
A potential challenge is label imbalance, which has been linked to example difficulty (\citealp{990132}; cf. \S\ref{adx:sup-res:dynasent original}). Therefore, we convert the dataset to a ternary label scheme by removing all datapoints that at least one annotator marked as `mixed,' then subsample the dataset to guarantee an equal distribution of positive, neutral and negative labels. 
%\footnote{
%    Refer to \Cref{adx:sup-res:dynasent original} for corresponding results on the original non-balanced DynaSent dataset.
%}
This re-balancing procedure severely limits the size of our dataset to roughly 32K training instances (see \S\ref{adx:deets:data}).
We report results on this re-balanced dataset, using the same PLM pools as in our previous experiments. 
Given the limited data, we train 1B pool models on the full dataset, and report results across 3 runs,  whereas we only train 1 seed for each of the 24 models from \citet{turc2019} in our homogeneous pool.
% To be written properly:
% \begin{itemize}
%     \item we're using the round 1 data from yelp exclusively
%     \item this contains 4 labels (positive, negative, neutral, and mixed), and 5 human judgments for each datapoint. We drop any datapoint where we don't have a consensus  among the annotators
%     \item given that label imbalance has been suggested as a factor of example difficulty \citep{990132}, but not as a factor of human label variation, we re-balance the dataset by (i) dropping anything with the mixed label (only 4\% of the training datapoints have a majority of ``mixed'' labels), as well as all datapoints where at least one of the annotator marked the item as `mixed', and then (ii) sub-sampling the data to have as many items in each of the categories. We refer to this subsampled dataset as the ``re-balanced'' dataset in the tables, as opposed to the ``original'' dataset which contains all the data available. 
%     \item we then consider two means of training classifiers: A, the usual standard approach of picking the majority among annotators as the gold label, and B, directly optimize the classifier to match its output distribution over labels with the human label distribution in dynasent. A is "hard labels", and B is "soft labels"
%     \item in total we have four distinct sub experiments: re-balanced dataset with soft labels, re-balanced dataset with hard labels, original dataset with soft labels,  original dataset with hard labels.
%     \item then somehow we had the brilliant idea to duplicate the work for 1B and <1B, so now there are 8 groups.
% \end{itemize}

\subsection{Results}

\input{Tables/dynasent-results/spearman-vs-h-ent-balanced}

\Cref{tab:dynasent:corrtable} summarizes the correlation between  model-based and human-based indicators on Dynasent. Due to the dataset's structure---specifically  the limited number of annotations per datapoint and their majority label selection---$\mathbb{H}_\mathrm{ent}$ and $\mathbb{H}_\mathrm{dis}$ are perfectly correlated.
We replicate observations made for NLI in \S\ref{sec:chaosNLI}:
When training the classifiers using hard labels, we observe a clear divide between reference-free and reference-dependent indicators.

Using soft labels yields obvious improvements for some of the reference-free indicators: 
In particular $\mathbb{M}_\mathrm{CP}$ and $\mathbb{M}_\mathrm{avg\ ent}$ are in some case competitive with reference-dependent metrics. 
It is however crucial to highlight that results remain volatile, as attested by the anti-correlations yielded by $\mathbb{M}_{\mathrm{CP\ }\alpha=0.2}$;\footnote{
    There are several possibility as to what causes this unexpected pattern for $\mathbb{M}_{\mathrm{CP\ }\alpha=0.2}$; we consider in particular the small size of the dataset as the behavior is not reproduced when training on the non-subsampled dataset, cf. \S\ref{adx:sup-res:dynasent original}.
} remark also that the optimal risk tolerance $\alpha$ we observe on DynaSent is $0.05$, instead of $0.1$ as we observed for NLI models.
Nor is the gap between reference-free and reference-dependent metrics fully mitigated: 
In all cases, the reference-dependent indicator $\mathbb{M}_{\mathrm{avg~ckpt~}p}^\mathrm{ref}$ outperforms all other indicators.
Taking stock of which indicators strongly benefit from soft-label trainings (viz. $\mathbb{M}_\mathrm{CP}$, $\mathbb{M}_\mathrm{avg\ ent}$ and $\mathbb{M}_{\mathrm{avg~ckpt~}p}^\mathrm{ref}$), we remark that they are derived from the probability distribution, rather than its argmax.
Soft labels foster distributions that are more in line with human label variation, but this might not suffice to fully bridge the gap between reference-free and reference-dependent indicators.\footnote{
    This is also in line with the fact that reference-free indicators derived from classifiers trained on soft labels still conflate success and failure, as shown in \S\ref{adx:sup-res:breakdown along success}, \Cref{tab:dynasent balanced:corrtable-models-partitioned}.
}
}

\section{Discussion}
Our study of how different indicators of data complexity correlate to one another has shown a somewhat perplexing picture worth diving into.
As we have established, model-based indicators align poorly with human-based indicators --- while we often observe positive correlations, their magnitudes are low. %, and linear regressions highlight significant residuals that are yet to be accounted for.
Defining model-based indicators with respect to human majority labels partially narrows the gap between the two, primarily because reference-free indicators often converge on a single label, regardless of its alignment with human preferences or the strength of consensus within the annotator pool. 
Within the reference-free indicators, we can also tentatively distinguish two subgroups: assessments that rely only on the pool of models considered (\cref{eq:model-ref dis,eq:model-ref pop ent}) appear to have a distinct profile from those which rely on more complex statistics, such as CP set sizes or entropy (\cref{eq:model-ref avg ent,eq:model-ref cp}).
For CP specifically, it is worth stressing that desiderata in terms of coverage can also entail significant variability. %, with higher values of $\alpha$ being less in line with human label variation in our experiments.

\textcolor{black}{Training classifiers to directly predict human label variation does not fully bridge the gap between reference-free and reference-dependent indicators, and only improves correlations with human assessment for indicators that do not summarize an model's distribution to its argmax.}
% In short, we find that m
Models often overwhelmingly agree on labels that lack humans annotator consensus, and factoring in human preferences in indicators is necessary though not sufficient for bridging the gap between human-based and model-based assessments difficulty. 
This underscores a critical limitation of the current research landscape: Reference-free approaches such as CP or entropy are at odds with reference-dependent approaches \citep[e.g.,][]{swayamdipta-etal-2020-dataset,baldock2021deep}, in that the former %have the critical flaw that they 
conflate failures and successes.\footnote{
    A related train of thought that can shed more light on our observations consists in considering which factors shape model decisions. See \S\ref{adx:sup-res:param counts} for a discussion. % of parameter counts, PLMs and fine-tuning data.
} % (a fact already hinted at by \citealp{swayamdipta-etal-2020-dataset}).

\textcolor{black}{Practical engineering recommendations also emerge from our observations.
Authors interested in developing automated assessments of data complexity in line with human assessments should favor (i) training models on soft labels, (ii) factoring in the actual probability distribution of the model, and (iii) leveraging the human label distribution, e.g., through the majority label. }

% This naturally leads us to studying what factors shape model dissensus.
% Interestingly, much of the variation in behavior across models seems to be due to parameter count, rather than pretraining and fine-tuning conditions, but much remains to be done to properly ascertain what are the most crucial factors at play. 
% Uncertainty disentanglement, as proposed by \citet{mucsanyi2024disentanglement}, offers a promising avenue for deeper exploration into these dynamics.  
% Critically, the effect sizes we associate with fine-tuning, pre-training and parameters appear noticeably small --- especially when compared against the remarkable stability that correlations within groups suggest, regardless of the pool of models considered. % whether we compare a BERT-based pool or a PLM pool of models.

In all, the present observations highlight a disconnect in the current literature. %push back on narratives such as \citeauthor{hu2023uncertainty}'s \citeyearpar{hu2023uncertainty}:
If data uncertainty is to be accounted for by factors such as noise, ambiguity or label overlap during data collection --- factors that we also expect to weigh in on measurements of linguistic disagreement --- then there is a need to reconcile this line of thought with the limited predictability of model-based assessments of data complexity from annotators' preferences.

\section{Conclusions}
We present a study with 11 indicators and 29 models, which show that human-based assessments of difficulty need not align with model-based assessments \citep{pavlick-kwiatkowski-2019-inherent} and that model-based assessments exhibit stark differences according to whether they factor in % a label grounded in 
human preferences.
%
% While our study is limited in that it only covers three datasets, this work is best construed as a counterexample: 
Data complexity and annotator disagreements, as assessed by model-based indicators or annotator label distribution,
have clearly distinct behaviors, %are at the very least distinct when it comes to NLI applications, 
despite the overlap the literature posits \citep{Lalor2018-ft}.
This calls for replication of our study in other settings, other tasks, other languages, etc.: %Here, we have identified a confound underlying discussions of data uncertainty and difficulty in NLP, but establishing its prevalence remains a topic for future work. 
Establishing the prevalence of the confound we identify remains a topic for future work. 

Lastly, our findings also question practices adopted by the field. 
If we are to posit a sharp distinction between data complexity as exemplified by \citet{swayamdipta-etal-2020-dataset} or \citet{baldock2021deep}, vs. uncertainty as captured by e.g. conformal prediction methods, then we need to explain why said data complexity is more in line with annotator disagreement than CP-based estimates of uncertainty.
Likewise, model-based estimates used in active learning \citep[e.g.][]{schroder-etal-2022-revisiting,baumler-etal-2023-examples} do not align with all definitions of uncertainty, especially label uncertainty as assessed through inter-annotator agreements. 
Such an exercise in terminology is a necessary step forward if we are to address challenges such as disentangling sources of uncertainty \citep{mucsanyi2024disentanglement} or leveraging uncertainty as a richer training signal \citep{basile-etal-2021-need, PalomakiRT18}.

%Potential application of our current study can be post analysis of data, difficulty with our proposed indicators and investigating its translation to another downstream tasks.
%
\makeatletter\ifacl@finalcopy
\section*{Acknowledgements}

This work was supported by the ICT 2023 project ``Uncertainty-aware neural language models'' funded by the Academy of Finland (grant agreement  \textnumero{}~345999). 

\fi
\makeatother

\section*{Limitations}
We identify two core limitations on our findings.

First, the present study relies on two datasets, namely the ChaosNLI re-annotation by \citet{nie-etal-2020-learn} and the DynaSent dataset of \citet{potts-etal-2021-dynasent}. 
While this limits the usefulness of our findings, and entails that our results might not carry on to other setups, we believe this choice is practically necessary (in that very few datasets are available with a training split large enough to easily train classifiers). It is in fact debatable whether DynaSent squarely meets all desiderata, since its validation split might not contain a rich enough set of annotations to accurately capture human label variation: DynaSent only collects five judgments from crowd workers, which \citet{nie-etal-2020-learn} shows to be unreliable.
On a practical level, this also means that there are many human-based indicators that we have ignored; e.g., the `complicated' label of \citet{jiang-marneffe-2022-investigating} or other explicit self-reports of uncertainty from the annotators could yield valuable insight that would contrast with the label distribution--based indicators we consider in \cref{eq:human dis,eq:human ent}.
Of course, all studies need to define their scope: In our case, more can always be done to integrate other data uncertainty/difficulty indicators from a wider range of studies, beyond the key ones we study here (viz. \citealp{nie-etal-2020-learn,vovk-etal-2005-algorithmic,baldock2021deep,swayamdipta-etal-2020-dataset}).

% Second, we have elected to emphasize descriptive statistics, rather than put a number on the predictive strength of our observations: 
% That is to say, most of this study focuses on describing what we observed on a dataset, and limited efforts have been made towards a statistical modeling and a quantification of the factors at play. 
% While there are justifications for this stance --- in particular, early experiments with mixed-effects linear models failed to reach convergence, owing to the limited correlations between the different series we consider ---  this does curtail the empirical strength of our approach, since we leave for future work all aspects related to understanding how the confound we identify will \textsl{a prima facie} bear weight on other works.
% Nonetheless, we firmly believe that the observations we report are worth taking into account.

Second, we rely on pool of models that %are trained with default hyperparameters, and 
have not been individually optimized for the task they are tested on. 
This point bears further discussion: As we identify in \Cref{tab:corrtable-models}, a major driver of the difference between reference-free and reference-dependent indicators is whether or not the classifier correctly identifies the gold label; and it therefore stands to reason that better trained classifiers may exhibit different patterns.
There are however three key facts to stress here.
First, hyperparameter tuning over a large pool of models (24 BERT variants from \citep{turc2019}, plus 25 1B PLMs--based classifiers, on three different datasets) is computationally prohibitive and would actively hinder the reproducibility of our experiments, which justifies the practice of limiting hyperparameter searches.
Second, our discussion pertains to the general usefulness of the indicators, rather than the fitness of the models --- or in other words, it is reasonable to expect of indicators of data complexity that they be robust enough to be deployed with less-than-top-of-the-leaderboard models.
Third, going by \Cref{tab:corrtable-models}, the main driver for the limited correlation between the different groups of model-based indicators is their failure, i.e.,  the main insight is that we would observe higher correlations if the models never failed, which is not a very realistic standard to expect from NLP systems.
While we believe this justifies our approach, it is quite plausible that the exact results as reported here would shift towards higher correlations with human assessments should the models reach higher accuracy scores.

%

\bibliography{anthology,custom}

\appendix

\section{Implementation details}
\label{adx:deets}

Our use of all preexisting research artifacts is consistent with their corresponding licenses.
We trust creators of said artifacts to have handled any personally identifying information that the artifacts may contain.

\textcolor{black}{We also provide our code for replication purposes at \href{https://anonymous.4open.science/r/data-cplx-unc-52CA/README.md}{this anonymized link}.}

\begin{table}[th]
    \centering
    \resizebox{0.95\columnwidth}{!}{
    \begin{tabular}{ll *{3}{S[table-format=6.0, round-precision=0, round-mode=places]}}
    \toprule
       \textbf{Dataset} & \textbf{Variant} & {\textbf{Train}} & {\textbf{Val}}  & {\textbf{Test}} \\
       \midrule
       \multirow{2}{*}{\textbf{SNLI}}
       & All labeled (\S\ref{sec:chaosNLI}, <1B) & 549367 & 9842 & {(\emph{unused})} \\
       & 5-splits (\S\ref{sec:chaosNLI}, 1B) & 109873 & 9842 & {(\emph{unused})} \\
        \midrule
       \multirow{2}{*}{\textbf{MNLI}}
       & All labeled (\S\ref{sec:chaosNLI}, <1B) & 392702 & 9815 & {(\emph{unused})} \\
       & 5-splits (\S\ref{sec:chaosNLI}, 1B) & 78540 & 9815 & {(\emph{unused})} \\
        \midrule
       \multirow{2}{*}{\textbf{DynaSent}}
       & Re-balanced (\S\ref{sec:dynasent}) & 32001 & 3066 & 3027 \\
       & All labeled (\S\ref{adx:sup-res:dynasent original}) & 84388 & 3600 & 3600  \\
        \bottomrule
    \end{tabular}
    }
    \caption{Dataset Statistics}
    \label{tab:dataset-stats}
\end{table}

\subsection{Data} 
\label{adx:deets:data}

As noted above, we use SNLI (\citealp{bowman-etal-2015-large}: retrieved from \href{https://huggingface.co/datasets/stanfordnlp/snli}{HuggingFace}), MNLI (\citealp{williams-etal-2018-broad}, retrieved from \href{https://huggingface.co/datasets/nyu-mll/multi_nli}{HuggingFace}), ChaosNLI (\citealp{nie-etal-2020-learn}; retrieved from \href{https://github.com/easonnie/ChaosNLI}{GitHub}) and DynaSent (\citealp{potts-etal-2021-dynasent}, retrieved from \href{https://github.com/cgpotts/dynasent}{GitHub}, round 1 data). 
We remove items without public labels from SNLI and MNLI, as well as datapoints with no majority label from DynaSent.
% The data therefore corresponds to 549367 training items for the <1B model pool,
% and either 109873 or 109874 training items for the 1B model pool.
Corresponding statistics are listed in \Cref{tab:dataset-stats}.

Data shuffling was seeded (with fixed random seeds per runs) for DynaSent experiments so as to guarantee strictly comparable training conditions between soft and hard label experiments. 

\subsection{Models} 

All models are implemented with HuggingFace (HF; \citealp{wolf-etal-2020-transformers,lhoest-etal-2021-datasets}). 
As per default HF implementations, for the 1B pool of models, classifiers rely on the last token in the input; for the <1B model pool, we use the first token. 
All experiments are supervised full fine-tuning processes using learned linear projections as classification heads.
Models are trained on a V100 NVIDIA GPU, for an individual runtime of $\leq$15 hours for any individual model.

All classifier heads for DynaSent were initialized (with fixed random seeds per run) so as to guarantee strictly comparable training conditions between soft and hard label experiments. 

\begin{table}[th]
    \centering
    \begin{subfigure}[b]{0.95\columnwidth}   \centering
    \begin{tabular}{>{\bf\footnotesize}p{0.7\linewidth} >{\RaggedLeft\arraybackslash\footnotesize}p{0.2\linewidth}} 
    \toprule
        Number of epochs & 2 \\
        Batch size & 16 \\
    \bottomrule
    \end{tabular}
    \caption{Hyperparameters, <1B SNLI models}
    \end{subfigure}
    
    \begin{subfigure}[b]{0.95\columnwidth}   \centering
    \begin{tabular}{>{\bf\footnotesize}p{0.7\linewidth} >{\RaggedLeft\arraybackslash\footnotesize}p{0.2\linewidth}}
    \toprule
        Number of epochs & 10 \\
        Batch size & 16 \\
    \bottomrule
    \end{tabular}
    \caption{Hyperparameters, 1B SNLI models}
    \end{subfigure}
    
    \begin{subfigure}[b]{0.95\columnwidth}   \centering
    \begin{tabular}{>{\bf\footnotesize}p{0.7\linewidth} >{\RaggedLeft\arraybackslash\footnotesize}p{0.2\linewidth}}
    \toprule
        Number of epochs & 5 \\
        Batch size & 1 \\
        Gradient accumulation & 16 \\
        Warmup ratio & 0.1 \\
    \bottomrule
    \end{tabular}
    \caption{Hyperparameters, all DynaSent models}
    \end{subfigure}
    
    \begin{subfigure}[b]{0.95\columnwidth}   \centering
    \begin{tabular}{>{\bf\footnotesize}p{0.7\linewidth} >{\RaggedLeft\arraybackslash\footnotesize}p{0.2\linewidth}}
    \toprule
        Number of epochs & 5 \\
        Batch size & 1 \\
        Gradient accumulation & 16 \\
        Warmup ratio & 0.1 \\
        Learning rate & 1e-6 \\
    \bottomrule
    \end{tabular}
    \caption{Hyperparameters, all MNLI models}
    \end{subfigure}

    \caption{Hyperparameters for all models considered}
    \label{tab:hyperparameters}
\end{table}

Hyperparameters are listed in \Cref{tab:hyperparameters}. Any hyperparameter not listed in \Cref{tab:hyperparameters} was left to its default value as listed in the HF documentation.
% As noted in the limitations section, no hyperparameter optimization was performed.

\color{black}
\section{Supplementary results}
\label{adx:sup-res}

\subsection{Non-linear relationship of human-based indicators and model-based indicators}
\label{adx:sup-res:residual analysis}

\input{Tables/r2table}

To get a better grasp on the magnitude of the difference highlighted in \Cref{tab:snli:corrtable,tab:mnli:corrtable}, we can turn to residual analyses.
We fit a linear regression, attempting to predict one indicator from another, and measure the proportion of variance that this linear model can explain using a coefficient of determination $R^2$.
Corresponding values are shown in \Cref{tab:r2}, with \Cref{tab:r2-bert} focusing on the <1B group and \Cref{tab:r2-1b} the 1B group.
In short, $R^2$ are never above $20\%$, and often below $10\%$ for reference-free metrics, suggesting that at least $80\%$ of the behavior of our model-based indicators cannot be accounted for with human-based indicators alone.
In this case as well, we can observe a difference between reference-free and reference-dependent indicators:
As one would expect, reference-dependent indicators yield quantitatively higher $R^2$ scores, suggesting they are (marginally) more in line with human indicators.
It is worth highlighting that out of all the reference-free indicators, conformal prediction set sizes and average model entropy scores tend to be the most in line with human judgments. This echoes our earlier remarks on the reference-free indicators being partitioned in two sub-groups, and suggests that more elaborate statistical estimators may mitigate some of the discrepancy we observe between human-based and model-based indicators.

\subsection{Interaction of model-based indicators and model success}
\label{adx:sup-res:breakdown along success}

A more formal statement of the argument shown in \Cref{fig:example-model-v-model} is that  we observe higher correlations when comparing two model-based indicators than when comparing human-based to model-based indicators, although correlations remain smaller than what we observe when comparing indicators within the same group. 
This can be seen in \Cref{tab:corrtable-models} for SNLI, where the values are clearly below what we observe within any subgroup of indicators, but higher than what we summarized in \Cref{tab:snli:corrtable}.

\input{Tables/corrtable-models}

To show that all of our reference-free indicators conflate model-success and failure, we can break  down observations depending on whether over $50\%$ of the model pool produces the majority label of the human annotator pools. 
Recomputing correlations for each subgroup yields systematic negative correlations when the models tend to fail, and systematic positive correlations when the models tend to succeed according to the majority labels, as summarized in \Cref{tab:snli:corrtable-models-partitioned,tab:mnli:corrtable-models-partitioned,tab:dynasent balanced:corrtable-models-partitioned} --- informally, correlations form the `left leg' of the inverted U-shape distributions, and anti-correlation the `right leg.'

\input{Tables/snli-results/corrs-model-when-fail}

\input{Tables/mnli-results/corrs-model-when-fail}

\input{Tables/dynasent-results/corrs-model-when-fail}

We have already mentioned the surprising anti-correlation of $\mathbb{M}_{\mathrm{CP\ }\alpha=0.2}$ on DynaSent within the 1B pool in the main body of the text, here we see that this unexpected behavior also impacts its joint distributions.
Another case worth highlighting concerns $\mathbb{M}_{\mathrm{CP\ }\alpha=0.05}$ on MNLI within the 1B pool: correlations and anti-correlations are of remarkably lower magnitudes, which suggests that this specific indicator is not in line with any of the reference-dependent indicators we consider. 
This is in line with our remarks that the exact setup considered always plays a 
More broadly, these observations largely confirm our claims in the main body of this article: 
We find that in most cases, the discrepancy between reference-free and reference-dependent indicators is due to the former conflating model success and model failure.

\subsection{Factors shaping model dissensus}
\label{adx:sup-res:param counts}

We can also leverage the different pools of models to assess how their factors of variation might impact data complexity metrics.
In particular, our heterogeneous group of 1B models was defined with respect to different PLMs and training subsets, and therefore we can measure whether pretraining conditions are more impactful than supervised fine-tuning data. 
In practice, we can measure how likely it is that two predictions for a specific datapoint will match, given that they were made by classifiers trained from the same model or by classifiers trained on the same training subset of SNLI.
This can be measured using common language effect sizes derived from a Mann-Whitney U test. 
Doing suggests a statistically significant effect from both splits and models ($p < 10^{-44}$) with a very small effect size ($f \approx 51.2\%$) on SNLI.
On MNLI, we find a somewhat stronger effect ($f=53.10\%, p < \epsilon$) when considering classifiers derived from the same PLM; as for the training data, it appears to yield the opposite effect, though with a much higher $p$-value ($f=49.77\%, p<10 ^{-3}$). 
For DynaSent, recall we have no sub-splits to experiment with; however we do find a positive effect when considering classifiers derived from the same PLMs ($f=53.63\%, p <\epsilon$).

Our homogeneous <1B pool %of BERT-based models 
also allows us to look into whether responses are more likely to differ for two models with a larger difference in number of parameters.
To test this, we can measure the likelihood of the parameter count difference being larger when the predictions differ using U tests.
Doing so, we can observe a common language effect size of $f=45.96\%$ on SNLI, $f=45.63\%$ on MNLI, and $f=45.69\%$ on DynaSent.
We can likewise observe a similar effect when focusing on our heterogeneous pool: we find a common language effect size of $f=48.90\%$ for SNLI, $f=45.63\%$ for MNLI, and $f=43.72\%$ on DynaSent.
In other words, predictions that match tend to come from models with more similar parameter counts.

\subsection{Replication of DynaSent experiments without re-balancing}
\label{adx:sup-res:dynasent original}

For the sake of exhaustiveness, we include experimental results derived all usable datapoints in DynaSent, i.e., without applying the label re-balancing step we detail in \S\ref{sec:dynasent}.
This entails several key differences. 
In particular, we now consider a classification with imbalanced labels among four possible classes (instead of three evenly split classes as per \S\ref{sec:dynasent}), but have access to more data to fit our classifiers. 
Also note that by construction \citet{potts-etal-2021-dynasent} exclude datapoints classified by a majority of annotators as `mixed' from the test sets, meaning there is a clear distributional shift between train and test.

\begin{table}[t!]
\centering

     \resizebox{0.95\columnwidth}{!}{
        \input{Tables/dynasent-original/spearman-vs-h-ent-original}
    }
    
    \caption{Spearman's $\rho$ between model-based indicators vs. human-based (entropy) indicator on DynaSent.}
    \label{tab:dynasent original:corrtable}
\end{table}

Comparisons between model-based and human-based indicators are shown in \Cref{tab:dynasent original:corrtable}.
The main point to be stressed is that we replicate the core findings stressed in the main body of the text, with two key differences.
First, we do not observe an anti-correlation for $\mathbb{M}_{\mathrm{CP\ }\alpha=0.2}$ within the 1B pool pof models.
Second, the use of soft labels does not appear to be beneficial to any indicator derived from the <1B pool of models.
Soft labels do remain useful to 1B models, and we still observe that soft labels are not sufficient for reference-free indicators to systematically bridge the gap separating them from reference-dependent indicators.

\input{Tables/dynasent-original/corrs-model-when-fail}

We further include evidence for the inverted U-shapes of the joint distributions of reference-free and reference-dependent indicators in \Cref{tab:dynasent original:corrtable-models-partitioned}. 
Remarks similar to what we already discussed in \S\ref{adx:sup-res:breakdown along success} can be made.

Overall, this supplementary experiment suggests an interesting perspective for future work: Label-imbalance does impact our indicators, as \citet{990132} suggest, but its exact effects appear to be contingent on the exact pool of models under consideration.

\end{document}

%% file: background_new.tex
Our study falls at the intersection of data complexity and uncertainty in NLP, particularly in relation to annotator disagreement, label variation, and the metrics used to evaluate them \citep{jiang-marneffe-2022-investigating, Uma2021, lorena2019howcomplex, baan2023uncertainty}. Though closely related, these concepts involve distinct challenges. %We will first define them and then discuss their relation.

\textbf{Data uncertainty} or aleatoric uncertainty describes the \textit{randomness or noise inherent to the data}  \citep{hu2023uncertainty}. This type of uncertainty is \textit{irreducible} and cannot be eliminated through model improvements or tuning \citep{KIUREGHIAN2009105,kendall2017uncertainties,Hüllermeier2021}. Sources of data uncertainty include noisy observations, overlapping classes, ground truth errors or inherent randomness.

\textbf{Annotator disagreement} is highlighted as \textit{a fundamental characteristic of linguistic data}, stemming from both  annotation noise and the inherent ambiguity of language \citep{plank-etal-2014-linguistically, Aroyo2015, pavlick-kwiatkowski-2019-inherent,fornaciari-etal-2021-beyond}. It often  correlates with \textit{label uncertainty}, where no single correct label exists. 
High-disagreement examples contain valuable signals for classifiers \citep{basile-etal-2021-need, PalomakiRT18}. 
However, disagreement does not necessarily indicate annotation noise---it may instead reflect genuine linguistic or contextual ambiguity.

\textbf{Data complexity} or data difficulty refers to the \textit{characteristics of a data sample that make classification inherently difficult}. It is related to the structural properties of the data, not to randomness or noise, as is the case with data uncertainty. Several factors contribute to data complexity: proximity to decision boundaries and class overlap%increase classification difficulty by making it harder for models to separate classes; 
, semantic indeterminacy%, which involves the inherent vagueness of language%, can lead to multiple plausible labels for a data point
, and task-specific challenges, such as requiring world knowledge %, further complicate classification 
\citep{plank-2022-problem, jiang-marneffe-2022-investigating}.

\textbf{Metrics for evaluating data uncertainty and complexity} include model confidence (probability of the predicted class), entropy of predicted probabilities (measuring classification uncertainty), and confidence calibration (aligning confidence with performance). These help assess label uncertainty caused by overlapping class boundaries \citep[e.g.,][]{geng-etal-2024-survey,ZHOU2022449,Xiao_Wang_2019}.
Training dynamics, such as how quickly a model learns to classify an example or the shape of the loss curve, further reveal the relative difficulty of datapoints \citep{swayamdipta-etal-2020-dataset,toneva2019Forgetting,baldock2021deep}.
These metrics offer tools to analyze the challenges of data complexity, yet they do not provide a nuanced perspective on the interplay between annotator disagreement and model uncertainty. 

\color{black}
\paragraph{Complexity, disagreement \& uncertainty.}

Data complexity, annotator disagreement, and data uncertainty are intertwined phenomena with similar root causes. 
For instance, while authors differ in their terminology, `overlapping classes' \citep{990132, Peterson_2019_ICCV, lorena2019howcomplex}, `absence of a single ground truth' \citep{Aroyo2015,baan-etal-2022-stop}, or `linguistic ambiguity' due to semantic and social factors \citep{plank-2022-problem} all refer to the fact that some datapoints can have different labels ---  which drives up complexity, disagreement and uncertainty. 

Hence, these three phenomena have been conflated in the literature. For example, uncertainty is often measured through label distribution entropy, used as a proxy for annotator disagreement \citep{zhang-etal-2022-survey, baumler-etal-2023-examples}. %This risks oversimplifying the relationship between annotator disagreement and data complexity. 
Similarly, \citet{Lalor2018-ft} has found alignment between human difficulty and model-assigned probability mass, suggesting that both perceive difficulty similarly. 
This viewpoint is also prevalent in active learning \citep[e.g.,][]{hachey-etal-2005-investigating} and attested less directly in quality estimation \citep[e.g.,][]{jamison-gurevych-2015-noise}.
Additionally, anecdotal evidence has been used to support this connection, especially in studies exploring the underlying causes of data complexity \citep{swayamdipta-etal-2020-dataset, baldock2021deep, rajpurkar-etal-2018-know}.

This perspective has also been employed as a working hypothesis. For example, \citet{pmlr-v80-weinshall18a} assume that knowledge distillation from a model can play the same role as humans in active learning scenarios, while \citet{beigman-beigman-klebanov-2009-learning} propose replacing annotator-based disagreement assessment with classifier proxies.
Authors adopting this approach often treat it as a simplifying initial assumption, and frequently include modeling work to better capture human variation \citep[e.g.][]{reidsma-carletta-2008-squibs} or discussions of the limitations of this working hypothesis (as in \citeauthor{beigman-beigman-klebanov-2009-learning}).

Treating these three phenomena as interchangeable oversimplifies their relationships. Disagreement often signals semantic complexity but can also stem from bias, expertise variance, or cultural differences \citep[e.g.,][]{jiang-marneffe-2022-investigating}. Similarly, uncertainty overlaps with complexity but also arises from noise that is not tied to structural data complexity \citep{kendall2017uncertainties}, while example difficulty has been linked to factors that are \textsl{a priori} not linguistic, such as class imbalance or distributional shifts  \citep{990132,Gawlikowski2023}. % For instance, a text with typos might be highly uncertain but not semantically complex. Conflating label uncertainty with broader data complexity obscures nuances in human interpretation and model behavior. % , e.g., a high-entropy label distribution may indicate true ambiguity or result from annotator subjectivity, rather than inherent task difficulty.
While some studies find overlap between human disagreement and model uncertainty \citep{swayamdipta-etal-2020-dataset, baldock2021deep}, others challenge this view \citep{reidsma-carletta-2008-squibs}. Our findings highlight the need to distinguish these concepts, as we show that complexity metrics do not map linearly to human assessments.

\color{black}

%% file: Tables/snli-results/snli-spearman.tex
\begin{table}[t!]
\centering
    % \begin{subfigure}[b]{0.475\columnwidth}
    %     \centering
\resizebox{0.9\columnwidth}{!}{
\begin{tabular}{l  S[table-format=2.4] S[table-format=2.4] @{\qquad} S[table-format=2.4] S[table-format=2.4]  } %r rrr r rrrr} % <================================ p{6cm}
\toprule
% {}  & \multicolumn{2}{c}{\texttt{human-based}} && \multicolumn{3}{c}{\texttt{ref-free model-based}}  && \multicolumn{4}{c}{\texttt{ref-dep model-based}} \\
% \cmidrule{2-3}
% \cmidrule{5-7}
% \cmidrule{9-12}
{} & \multicolumn{2}{c@{\quad}}{<1B pool} & \multicolumn{2}{c}{1B pool} \\
{} &  $\mathbb{H}_\mathrm{ent}$ &  $\mathbb{H}_\mathrm{dis}$ &  $\mathbb{H}_\mathrm{ent}$ &  $\mathbb{H}_\mathrm{dis}$  \\%&&  $\mathbb{M}_\mathrm{dis}$ &  $\mathbb{M}_\mathrm{ent}$ &  $\mathbb{M}_\mathrm{avg\ ent}$ &&  $\mathbb{M}_\mathrm{fail}^{\mathrm{ref}}$ &  $\mathbb{M}_\mathrm{1^{st}\ layer}^{\mathrm{ref}}$ &  $\mathbb{M}_\mathrm{1^{st}\ ckpt}^{\mathrm{ref}}$ &  $\mathbb{M}_\mathrm{avg\ ckpt}^{\mathrm{ref}}$ \\

\midrule
$\mathbb{M}_\mathrm{dis}$ & 0.244002 & 0.217881 & 0.194749 & 0.177177  \\
$\mathbb{M}_\mathrm{ent}$ & 0.278397 & 0.243338 & 0.218292 & 0.197000  \\
$\mathbb{M}_\mathrm{avg\ ent}$ & 0.390060 & 0.348964 & 0.281053 & 0.239845  \\
$\mathbb{M}_{\mathrm{CP\ }\alpha=0.05}$ & 0.373667 & 0.318577 & 0.276692 & 0.231541  \\
$\mathbb{M}_{\mathrm{CP\ }\alpha=0.1}$ & 0.376300 & 0.337911 & 0.281884 & 0.239299  \\
$\mathbb{M}_{\mathrm{CP\ }\alpha=0.2}$ & 0.324827 & 0.306394 & 0.248218 & 0.215699  \\
\midrule
$\mathbb{M}_\mathrm{fail}^{\mathrm{ref}}$ & 0.398967 & 0.395935 & 0.349701 & 0.333037  \\
$\mathbb{M}_\mathrm{1^{st}\ layer}^{\mathrm{ref}}$ & 0.371887 & 0.379588 & 0.362358 & 0.338693  \\
$\mathbb{M}_\mathrm{1^{st}\ ckpt}^{\mathrm{ref}}$ & 0.435674 & 0.424448 & 0.368188 & 0.344319  \\
$\mathbb{M}_\mathrm{avg\ ckpt}^{\mathrm{ref}}$ & 0.396941 & 0.390403 & 0.347681 & 0.327350  \\
$\mathbb{M}_{\mathrm{avg\ ckpt\ }p}^{\mathrm{ref}}$ & 0.438645 & 0.424106 & 0.367049 & 0.342756  \\
\bottomrule
\end{tabular}
}
\caption{Spearman correlation between human-based and model-based indicators on SNLI.}
\label{tab:snli:corrtable}
\end{table}
%\vspace{-2cm}

%% file: Tables/mnli-results/mnli-spearman.tex
\begin{table}[t!]
\centering
\resizebox{0.9\columnwidth}{!}{
\begin{tabular}{l  S[table-format=2.4] S[table-format=2.4] @{\qquad} S[table-format=2.4] S[table-format=2.4]  } %r rrr r rrrr} % <================================ p{6cm}
\toprule
% {}  & \multicolumn{2}{c}{\texttt{human-based}} && \multicolumn{3}{c}{\texttt{ref-free model-based}}  && \multicolumn{4}{c}{\texttt{ref-dep model-based}} \\
% \cmidrule{2-3}
% \cmidrule{5-7}
% \cmidrule{9-12}
{} & \multicolumn{2}{c@{\quad}}{<1B pool} & \multicolumn{2}{c}{1B pool} \\
{} &  $\mathbb{H}_\mathrm{ent}$ &  $\mathbb{H}_\mathrm{dis}$ &  $\mathbb{H}_\mathrm{ent}$ &  $\mathbb{H}_\mathrm{dis}$  \\
\midrule

% $\mathbb{H}_\mathrm{ent}$ & 1.000000 & 0.896358 \\
% $\mathbb{H}_\mathrm{dis}$ & 0.896358 & 1.000000 \\
$\mathbb{M}_\mathrm{dis}$ & -0.002232 & -0.004482 & 0.141883 & 0.107431 \\
$\mathbb{M}_\mathrm{ent}$ & 0.002331 & -0.001140 & 0.158733 & 0.120148 \\
$\mathbb{M}_\mathrm{avg\ ent}$ & -0.007652 & -0.015789 & 0.132883 & 0.109529 \\
$\mathbb{M}_{\mathrm{CP\ }\alpha=0.05}$ & 0.010060 & -0.008535 & 0.078811 & 0.042506 \\
$\mathbb{M}_{\mathrm{CP\ }\alpha=0.1}$ & -0.007260 & -0.017336 & 0.179797 & 0.116383 \\
$\mathbb{M}_{\mathrm{CP\ }\alpha=0.2}$ & -0.018384 & -0.023108 & 0.158052 & 0.093621 \\
\midrule
$\mathbb{M}_\mathrm{fail}^{\mathrm{ref}}$ & 0.117376 & 0.150791 & 0.172586 & 0.224618 \\
$\mathbb{M}_\mathrm{1^{st}\ layer}^{\mathrm{ref}}$ & 0.068166 & 0.096622 & 0.204044 & 0.251401 \\
$\mathbb{M}_\mathrm{1^{st}\ ckpt}^{\mathrm{ref}}$ & 0.113171 & 0.147928 & 0.182928 & 0.230737 \\
$\mathbb{M}_\mathrm{avg\ ckpt}^{\mathrm{ref}}$ & 0.116766 & 0.149847 & 0.176423 & 0.226072 \\
$\mathbb{M}_{\mathrm{avg\ ckpt\ }p}^{\mathrm{ref}}$ & 0.109385 & 0.143412 & 0.181264 & 0.230670 \\
\bottomrule
\end{tabular}
}
\caption{Spearman correlation between human-based and model-based indicators on MNLI.}
\label{tab:mnli:corrtable}
\end{table}

%% file: Tables/dynasent-results/spearman-vs-h-ent-balanced.tex
\begin{table}[t!]
\centering

     \resizebox{0.9\columnwidth}{!}{
\begin{tabular}{l  S[table-format=2.4] S[table-format=2.4] @{\qquad} S[table-format=2.4] S[table-format=2.4] } %r rrr r rrrr} % <================================ p{6cm}
\toprule
 & \multicolumn{2}{c @{\qquad}}{<1B pool} & \multicolumn{2}{c}{1B pool} \\
 & {{soft}} & {{hard}} & {{soft}} & {{hard}} \\
\midrule
$\mathbb{M}_\mathrm{dis}$ &  0.067947 & 0.065404 &  0.155306  & 0.143457   \\
$\mathbb{M}_\mathrm{ent}$ &  0.070805 & 0.063082 &  0.160556  & 0.146313   \\
$\mathbb{M}_\mathrm{avg\ ent}$ &  0.139525 & 0.126554 &  0.199574  & 0.120575   \\
$\mathbb{M}_{\mathrm{CP\ }\alpha=0.05}$ & 0.122310 & 0.107492 &  0.190427  & 0.138634   \\
$\mathbb{M}_{\mathrm{CP\ }\alpha=0.1}$ & 0.117745 & 0.099780 &  0.183588  & 0.134914   \\
$\mathbb{M}_{\mathrm{CP\ }\alpha=0.2}$ & 0.109955 & 0.081290 & -0.087108  &-0.126243   \\
\midrule
$\mathbb{M}_\mathrm{fail}^{\mathrm{ref}}$ & 0.128621 & 0.115634 &  0.185803  & 0.176365   \\
$\mathbb{M}_\mathrm{1^{st}\ layer}^{\mathrm{ref}}$ & 0.131855 & 0.123511 &  0.201608  & 0.192833   \\
$\mathbb{M}_\mathrm{1^{st}\ ckpt}^{\mathrm{ref}}$ &  0.131310 & 0.113666 &  0.201567  & 0.190652   \\
$\mathbb{M}_\mathrm{avg\ ckpt}^{\mathrm{ref}}$ &  0.126832 & 0.113302 &  0.186492  & 0.177996   \\
$\mathbb{M}_{\mathrm{avg\ ckpt\ }p}^{\mathrm{ref}}$ &  0.150418 & 0.135961 &  0.225674  & 0.189274   \\
\bottomrule
\end{tabular}
}
    
    \caption{Spearman's $\rho$ between model-based indicators vs. human-based (entropy) indicator on DynaSent.}
    \label{tab:dynasent:corrtable}
\end{table}

%% file: Tables/r2table.tex
\begin{table}[!t]
\centering
    \begin{subfigure}[b]{0.475\columnwidth}
        \centering
        \resizebox{\textwidth}{!}{
\begin{tabular}{l S[table-format=1.4] S[table-format=1.4] } %r rrr r rrrr} % <================================ p{6cm}
\toprule
% {}  & \multicolumn{2}{c}{\texttt{human-based}} && \multicolumn{3}{c}{\texttt{ref-free model-based}}  && \multicolumn{4}{c}{\texttt{ref-dep model-based}} \\
% \cmidrule{2-3}
% \cmidrule{5-7}
% \cmidrule{9-12}
{} &  $\mathbb{H}_\mathrm{dis}$ &  $\mathbb{H}_\mathrm{ent}$  \\%&&  $\mathbb{M}_\mathrm{dis}$ &  $\mathbb{M}_\mathrm{ent}$ &  $\mathbb{M}_\mathrm{avg\ ent}$ &&  $\mathbb{M}_\mathrm{fail}^{\mathrm{ref}}$ &  $\mathbb{M}_\mathrm{1^{st}\ layer}^{\mathrm{ref}}$ &  $\mathbb{M}_\mathrm{1^{st}\ ckpt}^{\mathrm{ref}}$ &  $\mathbb{M}_\mathrm{avg\ ckpt}^{\mathrm{ref}}$ \\

\midrule
$\mathbb{M}_\mathrm{dis}$ & 0.04747234306575854 & 0.0595368273550021 \\
$\mathbb{M}_\mathrm{ent}$ & 0.05921348914736824 & 0.07750491519829272 \\
$\mathbb{M}_\mathrm{avg\ ent}$ & 0.1217758707554526 & 0.1521471457429756 \\
$\mathbb{M}_{\mathrm{CP\ }\alpha=0.05}$ & 0.1014912195482921 & 0.1396268902186315 \\
$\mathbb{M}_{\mathrm{CP\ }\alpha=0.1}$ & 0.1141837322392022 & 0.1416015480435845 \\
$\mathbb{M}_{\mathrm{CP\ }\alpha=0.2}$ & 0.0938772915135655 & 0.10551259030511206 \\
\midrule
$\mathbb{M}_\mathrm{fail}^{\mathrm{ref}}$ & 0.15676480714978347 & 0.15917468662153678 \\
$\mathbb{M}_\mathrm{1^{st}\ layer}^{\mathrm{ref}}$ & 0.14408684144841533 & 0.1383000630979223 \\
$\mathbb{M}_\mathrm{1^{st}\ ckpt}^{\mathrm{ref}}$ & 0.1801565035106062 & 0.18981176047108939 \\
$\mathbb{M}_\mathrm{avg\ ckpt}^{\mathrm{ref}}$ & 0.15241486239334423 & 0.15756182951267483 \\
$\mathbb{M}_{\mathrm{avg\ ckpt\ }p}^{+\mathrm{ref}}$ & 0.1798661861659998 & 0.1924091552288929 \\
\bottomrule

\end{tabular}
}
\caption{<1B models}
\label{tab:r2-bert}
    \end{subfigure}    
    \begin{subfigure}[b]{0.475\columnwidth}
        \centering
        \resizebox{\textwidth}{!}{
\centering
\begin{tabular}{l S[table-format=1.4] S[table-format=1.4] } %rrrrrrr}
\toprule
 & $\mathbb{H}_\mathrm{dis}$ & $\mathbb{H}_\mathrm{ent}$  \\%& $\mathbb{M}_\mathrm{dis}$ & $\mathbb{M}_\mathrm{ent}$ & $\mathbb{M}_\mathrm{avg\ ent}$ & $\mathbb{M}_\mathrm{fail}^{\mathrm{ref}}$ & $\mathbb{M}_\mathrm{1^{st}\ layer}^{\mathrm{ref}}$ & $\mathbb{M}_\mathrm{1^{st}\ ckpt}^{\mathrm{ref}}$ & $\mathbb{M}_\mathrm{avg\ ckpt}^{\mathrm{ref}}$ \\
\midrule
$\mathbb{M}_\mathrm{dis}$ & 0.03139161542877611 & 0.037927068357759164 \\
$\mathbb{M}_\mathrm{avg\ ent}$ & 0.05752574995613624 & 0.07899086695488722 \\
$\mathbb{M}_\mathrm{ent}$ & 0.03880889456158054 & 0.04765128929296225 \\
$\mathbb{M}_{\mathrm{CP\ }\alpha=0.05}$ & 0.05361119272418913 & 0.0765586459847869 \\
$\mathbb{M}_{\mathrm{CP\ }\alpha=0.1}$ & 0.057263871881726214 & 0.0794583111002618 \\
$\mathbb{M}_{\mathrm{CP\ }\alpha=0.2}$ & 0.04652625216889361 & 0.061612145687850006 \\
\midrule
$\mathbb{M}_\mathrm{fail}^{\mathrm{ref}}$ & 0.11091352157761192 & 0.12229102138651815 \\
$\mathbb{M}_\mathrm{1^{st}\ layer}^{\mathrm{ref}}$ & 0.1147126306877495 & 0.13130365215110174 \\
$\mathbb{M}_\mathrm{1^{st}\ ckpt}^{\mathrm{ref}}$ & 0.11855566224740632 & 0.13556215253702497 \\
$\mathbb{M}_\mathrm{avg\ ckpt}^{\mathrm{ref}}$ & 0.10715831954198263 & 0.12088222456162945 \\
$\mathbb{M}_{\mathrm{avg\ ckpt\ }p}^{+\mathrm{ref}}$ & 0.11748197412940853 & 0.13472509769493457 \\
\bottomrule
\end{tabular}
        }
\caption{1B models}
\label{tab:r2-1b}
    \end{subfigure}    
\caption{Proportion of explained variance ($R^2$) of linear regressions predicting a model-based indicator from a human-based indicator.}
\label{tab:r2}
\end{table}
%\vspace{-2cm}

%% file: Tables/corrtable-models.tex
\begin{table}[!t]
\centering
    \begin{subfigure}[b]{0.95\columnwidth}
        \centering
        \resizebox{\textwidth}{!}{
\begin{tabular}{l  S[table-format=1.4] S[table-format=1.4]  S[table-format=1.4] S[table-format=1.4] S[table-format=1.4] } %r rrr r rrrr} % <================================ p{6cm}
\toprule
{} & $\mathbb{M}_\mathrm{fail}^{\mathrm{ref}}$ & $\mathbb{M}_\mathrm{1^{st}\ layer}^{\mathrm{ref}}$ & $\mathbb{M}_\mathrm{1^{st}\ ckpt}^{\mathrm{ref}}$ & $\mathbb{M}_\mathrm{avg\ ckpt}^{\mathrm{ref}}$ & $\mathbb{M}_{\mathrm{avg\ ckpt\ }p}^{\mathrm{ref}}$ \\ 
\midrule
$\mathbb{M}_\mathrm{dis}$ & 0.6189861268165726 & 0.5098473896698903 & 0.605266337396023 & 0.609105564034997 & 0.5985649061417667\\
$\mathbb{M}_\mathrm{ent}$ & 0.6212439136947254 & 0.5132877193939146 & 0.6082753266391165 & 0.6117477150418171 & 0.6034898715803866\\
$\mathbb{M}_\mathrm{avg\ ent}$ & 0.590402389488938 & 0.49730963573882 & 0.6221164649972188 & 0.5876074735736897 & 0.6427560295696796\\
$\mathbb{M}_{\mathrm{CP\ }\alpha=0.05}$ & 0.4602033924918084 & 0.3761694200818783 & 0.484508374501304 & 0.45847777071334533 & 0.5206219829652392\\
$\mathbb{M}_{\mathrm{CP\ }\alpha=0.1}$ & 0.4601097405551004 & 0.374771277215643 & 0.48807555510733286 & 0.4572180283326773 & 0.5044465345512179\\
$\mathbb{M}_{\mathrm{CP\ }\alpha=0.2}$ & 0.3545970868441389 & 0.31696826093940234 & 0.37466485574031494 & 0.35036302769217686 & 0.36225916709433037\\

\bottomrule
\end{tabular}
}
\caption{<1B models}
\label{tab:corrtable-models-bert}
    \end{subfigure}    

\vspace{0.1cm}
    
    \begin{subfigure}[b]{0.95\columnwidth}
        \centering
        \resizebox{\textwidth}{!}{
\centering
\begin{tabular}{l  S[table-format=1.4] S[table-format=1.4]  S[table-format=1.4] S[table-format=1.4] S[table-format=1.4] } %rrrrrrr}
\toprule
{} & $\mathbb{M}_\mathrm{fail}^{\mathrm{ref}}$ & $\mathbb{M}_\mathrm{1^{st}\ layer}^{\mathrm{ref}}$ & $\mathbb{M}_\mathrm{1^{st}\ ckpt}^{\mathrm{ref}}$ & $\mathbb{M}_\mathrm{avg\ ckpt}^{\mathrm{ref}}$ & $\mathbb{M}_{\mathrm{avg\ ckpt\ }p}^{\mathrm{ref}}$  \\
\midrule
$\mathbb{M}_\mathrm{dis}$ & 0.5153761883566349 & 0.4966014542398473 & 0.502850249061289 & 0.5035163808041141 & 0.5047642626775849\\
$\mathbb{M}_\mathrm{ent}$ & 0.5167914655136111 & 0.4984444670998636 & 0.5046603478193759 & 0.5072714883350755 & 0.5127099118269272\\
$\mathbb{M}_\mathrm{avg\ ent}$ & 0.5291583964824619 & 0.5418660780468404 & 0.5467896496088375 & 0.5291958544641427 & 0.5559662948232953\\
$\mathbb{M}_{\mathrm{CP\ }\alpha=0.05}$ & 0.48604302738262256 & 0.4958270504375366 & 0.4972185806478393 & 0.4915543459044353 & 0.5285969157122781\\
$\mathbb{M}_{\mathrm{CP\ }\alpha=0.1}$ & 0.5215970581035403 & 0.5289912548932993 & 0.5353388665358203 & 0.5240980409756174 & 0.5526959837745666\\
$\mathbb{M}_{\mathrm{CP\ }\alpha=0.2}$ & 0.5232050950601169 & 0.5337744402263092 & 0.5436616165354897 & 0.5187582628885268 & 0.5337611769269603\\
\bottomrule
\end{tabular}
        }
\caption{1B models}
\label{tab:corrtable-models-1b}
    \end{subfigure}    
\caption{Spearman correlation between reference-dependent and reference-free indicators.}
\label{tab:corrtable-models}
\end{table}
%\vspace{-2cm}SpearmanSpearman

%% file: Tables/snli-results/corrs-model-when-fail.tex
\begin{table*}[ht!]
\centering
    \begin{subfigure}[t]{0.475\linewidth}
        \centering
        \resizebox{\textwidth}{!}{
\begin{tabular}{l  S[table-format=2.4] S[table-format=2.4]  S[table-format=2.4] S[table-format=2.4]  S[table-format=2.4] } %r rrr r rrrr} % <================================ p{6cm}
\toprule
{} & $\mathbb{M}_\mathrm{fail}^{\mathrm{ref}}$ & $\mathbb{M}_\mathrm{1^{st}\ layer}^{\mathrm{ref}}$ & $\mathbb{M}_\mathrm{1^{st}\ ckpt}^{\mathrm{ref}}$ & $\mathbb{M}_\mathrm{avg\ ckpt}^{\mathrm{ref}}$ & $\mathbb{M}_{\mathrm{avg\ ckpt\ }p}^{\mathrm{ref}}$ \\
\midrule
$\mathbb{M}_\mathrm{dis}$ & -0.8508170713387564 & -0.8518359932652027 & -0.8461062459887859 & -0.8085290976914292 & -0.7630962342792861\\
$\mathbb{M}_\mathrm{ent}$ & -0.7932777990347453 & -0.7929502083062141 & -0.78632077193156 & -0.7550821165448501 & -0.7006838370616151\\
$\mathbb{M}_\mathrm{avg\ ent}$ & -0.596889642504745 & -0.5616991113433497 & -0.5389778880346432 & -0.5828154311312033 & -0.5941485563385027\\
$\mathbb{M}_{\mathrm{CP\ }\alpha=0.05}$ & -0.395794668109176 & -0.38755442342086915 & -0.3874200810636159 & -0.37358051927680574 & -0.35521645329958007\\
$\mathbb{M}_{\mathrm{CP\ }\alpha=0.1}$ & -0.4965477832446856 & -0.46399741426152663 & -0.4670005957476133 & -0.4833148079244341 & -0.48243230445667906\\
$\mathbb{M}_{\mathrm{CP\ }\alpha=0.2}$ & -0.4392357920123506 & -0.4013954701826735 & -0.3973925781036997 & -0.42974846283591095 & -0.44034938616974895\\
\bottomrule
\end{tabular}
}
\caption{<1B models, datapoints where most models fail}
\label{tab:corrtable-models-bert-fail}
    \end{subfigure}\quad
    \begin{subfigure}[t]{0.475\linewidth}
        \centering
        \resizebox{\textwidth}{!}{
\begin{tabular}{l  S[table-format=2.4] S[table-format=2.4]  S[table-format=2.4] S[table-format=2.4]  S[table-format=2.4] } %r rrr r rrrr} % <================================ p{6cm}
\toprule
{} & $\mathbb{M}_\mathrm{fail}^{\mathrm{ref}}$ & $\mathbb{M}_\mathrm{1^{st}\ layer}^{\mathrm{ref}}$ & $\mathbb{M}_\mathrm{1^{st}\ ckpt}^{\mathrm{ref}}$ & $\mathbb{M}_\mathrm{avg\ ckpt}^{\mathrm{ref}}$ & $\mathbb{M}_{\mathrm{avg\ ckpt\ }p}^{\mathrm{ref}}$ \\ 
\midrule
$\mathbb{M}_\mathrm{dis}$ & 0.9507423018392105 & 0.6920396436745644 & 0.9187041773958613 & 0.9213288170461678 & 0.885867322221493\\
$\mathbb{M}_\mathrm{ent}$ & 0.9488979921160743 & 0.6912312315961555 & 0.9176693564866069 & 0.9201499298125827 & 0.8873391308374752\\
$\mathbb{M}_\mathrm{avg\ ent}$ & 0.820244808859112 & 0.599841268930958 & 0.8759704414516686 & 0.8123056701887569 & 0.9370739771963427\\
$\mathbb{M}_{\mathrm{CP\ }\alpha=0.05}$ & 0.5833020850723017 & 0.40261102885121897 & 0.6342086949788669 & 0.5763457786432227 & 0.7053277236528628\\
$\mathbb{M}_{\mathrm{CP\ }\alpha=0.1}$ & 0.6237030779125001 & 0.4339300727450078 & 0.6774656157089634 & 0.615625598645598 & 0.7166177219258842\\
$\mathbb{M}_{\mathrm{CP\ }\alpha=0.2}$ & 0.49854939696487793 & 0.40157129189314766 & 0.5316641592793474 & 0.48662048600329105 & 0.5165370077809275\\
\bottomrule
\end{tabular}
}
\caption{<1B models, datapoints where most models succeed}
\label{tab:corrtable-models-bert-succeed}
    \end{subfigure} 

\vspace{0.1cm}

    \begin{subfigure}[t]{0.475\linewidth}
        \centering
        \resizebox{\textwidth}{!}{
\begin{tabular}{l  S[table-format=2.4] S[table-format=2.4]  S[table-format=2.4] S[table-format=2.4] S[table-format=2.4] } %r rrr r rrrr} % <================================ p{6cm}
\toprule
{} & $\mathbb{M}_\mathrm{fail}^{\mathrm{ref}}$ & $\mathbb{M}_\mathrm{1^{st}\ layer}^{\mathrm{ref}}$ & $\mathbb{M}_\mathrm{1^{st}\ ckpt}^{\mathrm{ref}}$ & $\mathbb{M}_\mathrm{avg\ ckpt}^{\mathrm{ref}}$ & $\mathbb{M}_{\mathrm{avg\ ckpt\ }p}^{\mathrm{ref}}$\\
\midrule
$\mathbb{M}_\mathrm{dis}$ & -0.776107996619319 & -0.7592574835008695 & -0.7736549898033511 & -0.7135851180602661 & -0.6838414506678357\\
$\mathbb{M}_\mathrm{ent}$ & -0.7130765533159207 & -0.7075291578654423 & -0.7173699214546921 & -0.6538924798347177 & -0.6139997066068105\\
$\mathbb{M}_\mathrm{avg\ ent}$ & -0.5615428419851041 & -0.5264151905708809 & -0.5283444504230372 & -0.5302538502846015 & -0.5110531087230158\\
$\mathbb{M}_{\mathrm{CP\ }\alpha=0.05}$ & -0.36701570801046135 & -0.36331654812881026 & -0.3514698304791617 & -0.3388936619031305 & -0.3036608987243743\\
$\mathbb{M}_{\mathrm{CP\ }\alpha=0.1}$ & -0.47606012129463354 & -0.456520108259136 & -0.45751157666691855 & -0.4453044281131862 & -0.4181839945703947\\
$\mathbb{M}_{\mathrm{CP\ }\alpha=0.2}$ & -0.6426860231851883 & -0.5835943601095165 & -0.5966804984213927 & -0.6155765131601448 & -0.6115945666544722\\
\bottomrule
\end{tabular}
}
\caption{1B models, datapoints where most models fail}
\label{tab:corrtable-models-1b-fail}
    \end{subfigure}\quad
    \begin{subfigure}[t]{0.475\linewidth}
        \centering
        \resizebox{\textwidth}{!}{
\begin{tabular}{l  S[table-format=2.4] S[table-format=2.4]  S[table-format=2.4] S[table-format=2.4] S[table-format=2.4] } %r rrr r rrrr} % <================================ p{6cm}
\toprule
{} & $\mathbb{M}_\mathrm{fail}^{\mathrm{ref}}$ & $\mathbb{M}_\mathrm{1^{st}\ layer}^{\mathrm{ref}}$ & $\mathbb{M}_\mathrm{1^{st}\ ckpt}^{\mathrm{ref}}$ & $\mathbb{M}_\mathrm{avg\ ckpt}^{\mathrm{ref}}$ & $\mathbb{M}_{\mathrm{avg\ ckpt\ }p}^{\mathrm{ref}}$  \\ \midrule
$\mathbb{M}_\mathrm{dis}$ & 0.953597644714019 & 0.8928098178897862 & 0.9159041233139131 & 0.9002646007218885 & 0.8934266079741909\\
$\mathbb{M}_\mathrm{ent}$ & 0.9464156826807936 & 0.8890729984687762 & 0.9106777410585433 & 0.8979774439398894 & 0.8962369555231695\\
$\mathbb{M}_\mathrm{avg\ ent}$ & 0.8803287659433822 & 0.895500001480897 & 0.9115738716653745 & 0.869355367143352 & 0.9314749761450709\\
$\mathbb{M}_{\mathrm{CP\ }\alpha=0.05}$ & 0.7748078280880732 & 0.7938547237717022 & 0.7971204011660153 & 0.7758987148129014 & 0.8546496882538488\\
$\mathbb{M}_{\mathrm{CP\ }\alpha=0.1}$ & 0.8546042197275769 & 0.8601389866569815 & 0.881613616490214 & 0.8490946142478989 & 0.9103434761969595\\
$\mathbb{M}_{\mathrm{CP\ }\alpha=0.2}$ & 0.8995557926175551 & 0.8981804559506075 & 0.9319656810337612 & 0.8799303154339231 & 0.9165786108886745\\
\bottomrule
\end{tabular}
}
\caption{1B models, datapoints where most models succeed}
\label{tab:corrtable-models-1b-succeed}
    \end{subfigure} 

\caption{Spearman correlation on SNLI data between reference-dependent and reference-free indicators of data complexity, broken down by average model success or failure.}
\label{tab:snli:corrtable-models-partitioned}
\end{table*}
%\vspace{-2cm}

%% file: Tables/mnli-results/corrs-model-when-fail.tex
\begin{table*}[ht!]
\centering
    \begin{subfigure}[t]{0.475\linewidth}
        \centering
        \resizebox{\textwidth}{!}{
\begin{tabular}{l  S[table-format=2.4] S[table-format=2.4]  S[table-format=2.4] S[table-format=2.4]  S[table-format=2.4] } %r rrr r rrrr} % <================================ p{6cm}
\toprule
{} & $\mathbb{M}_\mathrm{fail}^{\mathrm{ref}}$ & $\mathbb{M}_\mathrm{avg\ ckpt}^{\mathrm{ref}}$ & $\mathbb{M}_\mathrm{1^{st}\ ckpt}^{\mathrm{ref}}$ & $\mathbb{M}_\mathrm{1^{st}\ layer}^{\mathrm{ref}}$ & $\mathbb{M}_{\mathrm{avg\ ckpt\ }p}^{\mathrm{ref}}$ \\
\midrule
$\mathbb{M}_\mathrm{dis}$ & -0.7675725684583323 & -0.7624823393150436 & -0.7812004238275361 & -0.6911357469590993 & -0.7379878512581651\\
$\mathbb{M}_\mathrm{ent}$ & -0.7322471123797256 & -0.7301077623776069 & -0.7469400921405462 & -0.6486761267858221 & -0.7034426056201022\\
$\mathbb{M}_\mathrm{avg\ ent}$ & -0.5241471615243317 & -0.5129969011302901 & -0.5147697825535686 & -0.5188824812848594 & -0.5043783521067885\\
$\mathbb{M}_{\mathrm{CP\ }\alpha=0.05}$ & -0.3462616693459686 & -0.3307063428018463 & -0.34837763648534364 & -0.3288795600059665 & -0.33199821505012156\\
$\mathbb{M}_{\mathrm{CP\ }\alpha=0.1}$ & -0.4078671068480368 & -0.3929819265493521 & -0.40834189091478723 & -0.396415906762425 & -0.3929576608419417\\
$\mathbb{M}_{\mathrm{CP\ }\alpha=0.2}$ & -0.5069688872024931 & -0.49491494196791314 & -0.5012543604463965 & -0.5041402037729162 & -0.49145195205107234\\
\bottomrule
\end{tabular}
}
\caption{<1B models, datapoints where most models fail}
\label{tab:mnli:corrtable-models-bert-fail}
    \end{subfigure}\quad
    \begin{subfigure}[t]{0.475\linewidth}
        \centering
        \resizebox{\textwidth}{!}{
\begin{tabular}{l  S[table-format=2.4] S[table-format=2.4]  S[table-format=2.4] S[table-format=2.4]  S[table-format=2.4] } %r rrr r rrrr} % <================================ p{6cm}
\toprule
{} & $\mathbb{M}_\mathrm{fail}^{\mathrm{ref}}$ & $\mathbb{M}_\mathrm{avg\ ckpt}^{\mathrm{ref}}$ & $\mathbb{M}_\mathrm{1^{st}\ ckpt}^{\mathrm{ref}}$ & $\mathbb{M}_\mathrm{1^{st}\ layer}^{\mathrm{ref}}$ & $\mathbb{M}_{\mathrm{avg\ ckpt\ }p}^{\mathrm{ref}}$  \\ 
\midrule
$\mathbb{M}_\mathrm{dis}$ & 0.9746941120706925 & 0.8188523044826821 & 0.9592771997197398 & 0.9275661426409757 & 0.9344996399599854\\
$\mathbb{M}_\mathrm{ent}$ & 0.9622573837763254 & 0.8053107729978002 & 0.9483819519696569 & 0.9293566807956555 & 0.9280918722407833\\
$\mathbb{M}_\mathrm{avg\ ent}$ & 0.7828355250642166 & 0.7235767197821017 & 0.800003411721415 & 0.9118404537257533 & 0.7529745398582173\\
$\mathbb{M}_{\mathrm{CP\ }\alpha=0.05}$ & 0.5769752946597344 & 0.581901978226195 & 0.5857457555331771 & 0.7119613047174637 & 0.5664193676561784\\
$\mathbb{M}_{\mathrm{CP\ }\alpha=0.1}$ & 0.6581407049147187 & 0.62815121918385 & 0.6679961527786332 & 0.79707635175579 & 0.6466256368226817\\
$\mathbb{M}_{\mathrm{CP\ }\alpha=0.2}$ & 0.7559110165844621 & 0.6752949550713748 & 0.7693013009225297 & 0.8887532394975292 & 0.73857080795787\\
\bottomrule
\end{tabular}
}
\caption{<1B models, datapoints where most models succeed}
\label{tab:mnli:corrtable-models-bert-succeed}
    \end{subfigure} 

\vspace{0.1cm}

    \begin{subfigure}[t]{0.475\linewidth}
        \centering
        \resizebox{\textwidth}{!}{
\begin{tabular}{l  S[table-format=2.4] S[table-format=2.4]  S[table-format=2.4] S[table-format=2.4] S[table-format=2.4] } %r rrr r rrrr} % <================================ p{6cm}
\toprule
{} & $\mathbb{M}_\mathrm{fail}^{\mathrm{ref}}$ & $\mathbb{M}_\mathrm{1^{st}\ layer}^{\mathrm{ref}}$ & $\mathbb{M}_\mathrm{1^{st}\ ckpt}^{\mathrm{ref}}$ & $\mathbb{M}_\mathrm{avg\ ckpt}^{\mathrm{ref}}$ & $\mathbb{M}_{\mathrm{avg\ ckpt\ }p}^{\mathrm{ref}}$ \\
\midrule
$\mathbb{M}_\mathrm{dis}$ & -0.8043548195902573 & -0.7910007762147501 & -0.8041015497710354 & -0.7773240994039758 & -0.7784497532768622\\
$\mathbb{M}_\mathrm{ent}$ & -0.7593962529588667 & -0.7456938680642058 & -0.7640232367577571 & -0.7346455582194128 & -0.7320164990343113\\
$\mathbb{M}_\mathrm{avg\ ent}$ & -0.5422102023079934 & -0.4932810355916139 & -0.4884164099742378 & -0.5258558424681306 & -0.5364847565788831\\
$\mathbb{M}_{\mathrm{CP\ }\alpha=0.05}$ & -0.14152305627940642 & -0.11855946526349465 & -0.16142892969345837 & -0.12461880643196427 & -0.11057483548722075\\
$\mathbb{M}_{\mathrm{CP\ }\alpha=0.1}$ & -0.39126832578717063 & -0.36096216419364746 & -0.4014032858775856 & -0.3724467378492834 & -0.37058336359181937\\
$\mathbb{M}_{\mathrm{CP\ }\alpha=0.2}$ & -0.5293176282971108 & -0.48428642258358895 & -0.5179104808053508 & -0.5154807000884867 & -0.5181567385057368\\
\bottomrule
\end{tabular}
}
\caption{1B models, datapoints where most models fail}
\label{tab:mnli:corrtable-models-1b-fail}
    \end{subfigure}\quad
    \begin{subfigure}[t]{0.475\linewidth}
        \centering
        \resizebox{\textwidth}{!}{
\begin{tabular}{l  S[table-format=2.4] S[table-format=2.4]  S[table-format=2.4] S[table-format=2.4] S[table-format=2.4] } %r rrr r rrrr} % <================================ p{6cm}
\toprule
{} & $\mathbb{M}_\mathrm{fail}^{\mathrm{ref}}$ & $\mathbb{M}_\mathrm{1^{st}\ layer}^{\mathrm{ref}}$ & $\mathbb{M}_\mathrm{1^{st}\ ckpt}^{\mathrm{ref}}$ & $\mathbb{M}_\mathrm{avg\ ckpt}^{\mathrm{ref}}$ & $\mathbb{M}_{\mathrm{avg\ ckpt\ }p}^{\mathrm{ref}}$ \\ \midrule
$\mathbb{M}_\mathrm{dis}$ & 0.9766910654054102 & 0.8950873776267412 & 0.9413376084063438 & 0.9446982275928817 & 0.945856529736478\\
$\mathbb{M}_\mathrm{ent}$ & 0.9634957545025367 & 0.8875439060482998 & 0.9309399397477517 & 0.9337201011139045 & 0.9357645825328921\\
$\mathbb{M}_\mathrm{avg\ ent}$ & 0.7414275437042168 & 0.7344802214031794 & 0.7863200472608148 & 0.7338846471887537 & 0.7531997443338839\\
$\mathbb{M}_{\mathrm{CP\ }\alpha=0.05}$ & 0.05471712072564894 & 0.05396312210003065 & 0.03660296428744603 & 0.043222177690850724 & 0.03295557547460776\\
$\mathbb{M}_{\mathrm{CP\ }\alpha=0.1}$ & 0.5057723290031017 & 0.5203749090289149 & 0.5220069589082307 & 0.49948686641447687 & 0.5158946217782792\\
$\mathbb{M}_{\mathrm{CP\ }\alpha=0.2}$ & 0.6734400351650168 & 0.6805740689298293 & 0.709125045494041 & 0.6661026255722878 & 0.6936015972820667\\
\bottomrule
\end{tabular}
}
\caption{1B models, datapoints where most models succeed}
\label{tab:mnli:corrtable-models-1b-succeed}
    \end{subfigure} 

\caption{Spearman correlation on MNLI data between reference-dependent and reference-free indicators of data complexity, broken down by average model success or failure.}
\label{tab:mnli:corrtable-models-partitioned}
\end{table*}
%\vspace{-2cm}

%% file: Tables/dynasent-results/corrs-model-when-fail.tex
\begin{table*}[ht!]
\centering
    \begin{subfigure}[t]{0.475\linewidth}
        \centering
        \resizebox{\textwidth}{!}{
\begin{tabular}{l  S[table-format=2.4] S[table-format=2.4]  S[table-format=2.4] S[table-format=2.4]  S[table-format=2.4] } %r rrr r rrrr} % <================================ p{6cm}
\toprule
{} & $\mathbb{M}_\mathrm{fail}^{\mathrm{ref}}$ & $\mathbb{M}_\mathrm{1^{st}\ layer}^{\mathrm{ref}}$ & $\mathbb{M}_\mathrm{1^{st}\ ckpt}^{\mathrm{ref}}$ & $\mathbb{M}_\mathrm{avg\ ckpt}^{\mathrm{ref}}$ & $\mathbb{M}_{\mathrm{avg\ ckpt\ }p}^{\mathrm{ref}}$ \\
\midrule 
$\mathbb{M}_\mathrm{dis}$ & -0.7329097922961743 & -0.68916287490608 & -0.7322790773779071 & -0.6988534005922082 & -0.6435141328785403\\
$\mathbb{M}_\mathrm{ent}$ & -0.6539093540020625 & -0.6259152814538856 & -0.6519270975909313 & -0.6287451609202276 & -0.5508734470774038\\
$\mathbb{M}_\mathrm{avg\ ent}$ & -0.21060738638886267 & -0.2093873442443353 & -0.19255599495054587 & -0.19690268278171583 & -0.20938546205603942\\
$\mathbb{M}_{\mathrm{CP\ }\alpha=0.05}$ & -0.2055714629398085 & -0.22130874776967724 & -0.19030545936385404 & -0.20893191490083673 & -0.14632613953112344\\
$\mathbb{M}_{\mathrm{CP\ }\alpha=0.1}$ & -0.24493130473220648 & -0.25949423329225546 & -0.22484666746258702 & -0.24536555816972774 & -0.2066797895280547\\
$\mathbb{M}_{\mathrm{CP\ }\alpha=0.2}$ & -0.3729638458591846 & -0.3801358611644823 & -0.3464738651327859 & -0.3658177586884768 & -0.37426774087487064\\
\bottomrule
\end{tabular}
}
\caption{<1B models, soft labels, datapoints where most models fail}
\label{tab:dynasent balanced:soft:corrtable-models-bert-fail}
    \end{subfigure}\quad
    \begin{subfigure}[t]{0.475\linewidth}
        \centering
        \resizebox{\textwidth}{!}{
\begin{tabular}{l  S[table-format=2.4] S[table-format=2.4]  S[table-format=2.4] S[table-format=2.4]  S[table-format=2.4] } %r rrr r rrrr} % <================================ p{6cm}
\toprule
{} & $\mathbb{M}_\mathrm{fail}^{\mathrm{ref}}$ & $\mathbb{M}_\mathrm{1^{st}\ layer}^{\mathrm{ref}}$ & $\mathbb{M}_\mathrm{1^{st}\ ckpt}^{\mathrm{ref}}$ & $\mathbb{M}_\mathrm{avg\ ckpt}^{\mathrm{ref}}$ & $\mathbb{M}_{\mathrm{avg\ ckpt\ }p}^{\mathrm{ref}}$  \\ 
\midrule
$\mathbb{M}_\mathrm{dis}$ & 0.9777139239754828 & 0.9115315719054523 & 0.9665934339094299 & 0.9452224375619017 & 0.7375060171956368\\
$\mathbb{M}_\mathrm{ent}$ & 0.9501949181226401 & 0.9038985375475247 & 0.9406013399785405 & 0.9231399958838323 & 0.7441788022269105\\
$\mathbb{M}_\mathrm{avg\ ent}$ & 0.4917422830615071 & 0.5036649698385767 & 0.5106866613086157 & 0.45921080305769557 & 0.9306163214804769\\
$\mathbb{M}_{\mathrm{CP\ }\alpha=0.05}$ & 0.5433271662132608 & 0.5904258124744332 & 0.5592128387060755 & 0.5280467189755704 & 0.8927186573992236\\
$\mathbb{M}_{\mathrm{CP\ }\alpha=0.1}$ & 0.6237123657775733 & 0.6439450985565256 & 0.6423400029641175 & 0.6004670391900867 & 0.9367130750944331\\
$\mathbb{M}_{\mathrm{CP\ }\alpha=0.2}$ & 0.7350971682435353 & 0.7151018793929134 & 0.7541739171970422 & 0.7032017289208458 & 0.8974738066804604\\
\bottomrule
\end{tabular}
}
\caption{<1B models, soft labels, datapoints where most models succeed}
\label{tab:dynasent balanced:soft:corrtable-models-bert-succeed}
    \end{subfigure} 

\vspace{0.1cm}

    \begin{subfigure}[t]{0.475\linewidth}
        \centering
        \resizebox{\textwidth}{!}{
\begin{tabular}{l  S[table-format=2.4] S[table-format=2.4]  S[table-format=2.4] S[table-format=2.4]  S[table-format=2.4] } %r rrr r rrrr} % <================================ p{6cm}
\toprule
{} & $\mathbb{M}_\mathrm{fail}^{\mathrm{ref}}$ & $\mathbb{M}_\mathrm{1^{st}\ layer}^{\mathrm{ref}}$ & $\mathbb{M}_\mathrm{1^{st}\ ckpt}^{\mathrm{ref}}$ & $\mathbb{M}_\mathrm{avg\ ckpt}^{\mathrm{ref}}$ & $\mathbb{M}_{\mathrm{avg\ ckpt\ }p}^{\mathrm{ref}}$ \\
\midrule 
$\mathbb{M}_\mathrm{dis}$ & -0.7419798744343022 & -0.7208063225564367 & -0.743709800157977 & -0.7177971348223375 & -0.630957061670016\\
$\mathbb{M}_\mathrm{ent}$ & -0.679176208884438 & -0.6704652448344265 & -0.6803992534312436 & -0.6559486619211535 & -0.560942225545257\\
$\mathbb{M}_\mathrm{avg\ ent}$ & -0.32127736854770766 & -0.33497607448698874 & -0.3053469455523746 & -0.29779415940518905 & -0.2990667338021706\\
$\mathbb{M}_{\mathrm{CP\ }\alpha=0.05}$ & -0.3062227124056235 & -0.3126464546948456 & -0.29059746085892174 & -0.29449456549742375 & -0.24324250166095276\\
$\mathbb{M}_{\mathrm{CP\ }\alpha=0.1}$ & -0.38481749684167915 & -0.38052093448497215 & -0.36640785516347246 & -0.3747029782353623 & -0.34645334007764433\\
$\mathbb{M}_{\mathrm{CP\ }\alpha=0.2}$ & -0.4875360680354146 & -0.4821436782540145 & -0.4647314876656557 & -0.47930419243910827 & -0.47711430436559604\\\bottomrule
\end{tabular}
}
\caption{<1B models, hard labels, datapoints where most models fail}
\label{tab:dynasent balanced:hard:corrtable-models-bert-fail}
    \end{subfigure}\quad
    \begin{subfigure}[t]{0.475\linewidth}
        \centering
        \resizebox{\textwidth}{!}{
\begin{tabular}{l  S[table-format=2.4] S[table-format=2.4]  S[table-format=2.4] S[table-format=2.4]  S[table-format=2.4] } %r rrr r rrrr} % <================================ p{6cm}
\toprule
{} & $\mathbb{M}_\mathrm{fail}^{\mathrm{ref}}$ & $\mathbb{M}_\mathrm{1^{st}\ layer}^{\mathrm{ref}}$ & $\mathbb{M}_\mathrm{1^{st}\ ckpt}^{\mathrm{ref}}$ & $\mathbb{M}_\mathrm{avg\ ckpt}^{\mathrm{ref}}$ & $\mathbb{M}_{\mathrm{avg\ ckpt\ }p}^{\mathrm{ref}}$  \\ 
\midrule
$\mathbb{M}_\mathrm{dis}$ & 0.9832184616416267 & 0.8658862446709601 & 0.9580213438346206 & 0.9453909104994747 & 0.6923778680484735\\
$\mathbb{M}_\mathrm{ent}$ & 0.9692007337721718 & 0.8547211345570057 & 0.9447559432070015 & 0.9337482151924006 & 0.7022145566433206\\
$\mathbb{M}_\mathrm{avg\ ent}$ & 0.4627967944060504 & 0.44382832365128994 & 0.4834575043774163 & 0.46015489949039345 & 0.9264738169229724\\
$\mathbb{M}_{\mathrm{CP\ }\alpha=0.05}$ & 0.4725192502115214 & 0.4378850721861606 & 0.4943080058492018 & 0.48367843354790885 & 0.8639099733170399\\
$\mathbb{M}_{\mathrm{CP\ }\alpha=0.1}$ & 0.611556707247636 & 0.5831079337092638 & 0.6318114043140406 & 0.6129499508356914 & 0.9209144023805745\\
$\mathbb{M}_{\mathrm{CP\ }\alpha=0.2}$ & 0.7380282056234754 & 0.7528185226889765 & 0.7509227530718069 & 0.7194074538429687 & 0.8123518459946669\\
\bottomrule
\end{tabular}
}
\caption{<1B models, hard labels, datapoints where most models succeed}
\label{tab:dynasent balanced:hard:corrtable-models-bert-succeed}
    \end{subfigure} 

\vspace{0.1cm}

    \begin{subfigure}[t]{0.475\linewidth}
        \centering
        \resizebox{\textwidth}{!}{
\begin{tabular}{l  S[table-format=2.4] S[table-format=2.4]  S[table-format=2.4] S[table-format=2.4]  S[table-format=2.4] } %r rrr r rrrr} % <================================ p{6cm}
\toprule
{} & $\mathbb{M}_\mathrm{fail}^{\mathrm{ref}}$ & $\mathbb{M}_\mathrm{1^{st}\ layer}^{\mathrm{ref}}$ & $\mathbb{M}_\mathrm{1^{st}\ ckpt}^{\mathrm{ref}}$ & $\mathbb{M}_\mathrm{avg\ ckpt}^{\mathrm{ref}}$ & $\mathbb{M}_{\mathrm{avg\ ckpt\ }p}^{\mathrm{ref}}$ \\
\midrule 
$\mathbb{M}_\mathrm{dis}$ & -0.8089268130014203 & -0.7620147358884525 & -0.7934380854959021 & -0.7572492220842214 & -0.7498706246259447\\
$\mathbb{M}_\mathrm{ent}$ & -0.7166822303578448 & -0.6896891433503619 & -0.7126742008149825 & -0.678479941043295 & -0.6616065776103495\\
$\mathbb{M}_\mathrm{avg\ ent}$ & -0.42011769974791957 & -0.33646377302569436 & -0.3031311936337885 & -0.42807707784710286 & -0.4268185372554466\\
$\mathbb{M}_{\mathrm{CP\ }\alpha=0.05}$ & -0.4267769464658052 & -0.35789097102989825 & -0.35549557135854604 & -0.4237551743758893 & -0.4122387203785865\\
$\mathbb{M}_{\mathrm{CP\ }\alpha=0.1}$ & -0.43661040635572584 & -0.35939800018041346 & -0.34070175542786285 & -0.44701248280450495 & -0.4430003938858638\\
$\mathbb{M}_{\mathrm{CP\ }\alpha=0.2}$ & 0.3923893501308052 & 0.3417138800690526 & 0.29896559222123337 & 0.400585625877755 & 0.3995821018404135\\
\bottomrule
\end{tabular}
}
\caption{1B models, soft labels, datapoints where most models fail}
\label{tab:dynasent balanced:soft:corrtable-models-1b-fail}
    \end{subfigure}\quad
    \begin{subfigure}[t]{0.475\linewidth}
        \centering
        \resizebox{\textwidth}{!}{
\begin{tabular}{l  S[table-format=2.4] S[table-format=2.4]  S[table-format=2.4] S[table-format=2.4]  S[table-format=2.4] } %r rrr r rrrr} % <================================ p{6cm}
\toprule
{} & $\mathbb{M}_\mathrm{fail}^{\mathrm{ref}}$ & $\mathbb{M}_\mathrm{1^{st}\ layer}^{\mathrm{ref}}$ & $\mathbb{M}_\mathrm{1^{st}\ ckpt}^{\mathrm{ref}}$ & $\mathbb{M}_\mathrm{avg\ ckpt}^{\mathrm{ref}}$ & $\mathbb{M}_{\mathrm{avg\ ckpt\ }p}^{\mathrm{ref}}$  \\ 
\midrule
$\mathbb{M}_\mathrm{dis}$ & 0.8681102462637843 & 0.7185151159258553 & 0.8442458602089681 & 0.8292477260671024 & 0.7854488193127688\\
$\mathbb{M}_\mathrm{ent}$ & 0.8666413566027706 & 0.7169592892006007 & 0.8425284736123447 & 0.8278585071387423 & 0.784896465102505\\
$\mathbb{M}_\mathrm{avg\ ent}$ & 0.7937221265751566 & 0.7843357953863489 & 0.8145940087129888 & 0.7942701891323958 & 0.939870160448053\\
$\mathbb{M}_{\mathrm{CP\ }\alpha=0.05}$ & 0.7934643664767501 & 0.7714954801443785 & 0.8166055782915699 & 0.7922949662906035 & 0.9238700617857031\\
$\mathbb{M}_{\mathrm{CP\ }\alpha=0.1}$ & 0.8056097346506506 & 0.7311433035914203 & 0.8327966534503234 & 0.7980269634700562 & 0.868751170240836\\
$\mathbb{M}_{\mathrm{CP\ }\alpha=0.2}$ & -0.3794216733596527 & -0.36125463697693083 & -0.38803360848397206 & -0.37147857306789434 & -0.40325430000293844\\\bottomrule
\end{tabular}
}
\caption{1B models, soft labels, datapoints where most models succeed}
\label{tab:dynasent balanced:soft:corrtable-models-1b-succeed}
    \end{subfigure}

\vspace{0.1cm}

    \begin{subfigure}[t]{0.475\linewidth}
        \centering
        \resizebox{\textwidth}{!}{
\begin{tabular}{l  S[table-format=2.4] S[table-format=2.4]  S[table-format=2.4] S[table-format=2.4]  S[table-format=2.4] } %r rrr r rrrr} % <================================ p{6cm}
\toprule
{} & $\mathbb{M}_\mathrm{fail}^{\mathrm{ref}}$ & $\mathbb{M}_\mathrm{1^{st}\ layer}^{\mathrm{ref}}$ & $\mathbb{M}_\mathrm{1^{st}\ ckpt}^{\mathrm{ref}}$ & $\mathbb{M}_\mathrm{avg\ ckpt}^{\mathrm{ref}}$ & $\mathbb{M}_{\mathrm{avg\ ckpt\ }p}^{\mathrm{ref}}$ \\
\midrule 
$\mathbb{M}_\mathrm{dis}$ & -0.8120739843295619 & -0.7821776326943135 & -0.8183077049461951 & -0.7465583501121253 & -0.548102413841498\\
$\mathbb{M}_\mathrm{ent}$ & -0.7232669397629409 & -0.7044091055342431 & -0.7384448261819235 & -0.6693219997668488 & -0.4481417509067416\\
$\mathbb{M}_\mathrm{avg\ ent}$ & -0.4583930098601233 & -0.4193141506493951 & -0.41175555317593543 & -0.4363753848082686 & -0.2267176190620782\\
$\mathbb{M}_{\mathrm{CP\ }\alpha=0.05}$ & -0.46927309665670075 & -0.4369122222694643 & -0.40264848349977156 & -0.4592803808643976 & -0.3371437587988966\\
$\mathbb{M}_{\mathrm{CP\ }\alpha=0.1}$ & -0.46780477666916104 & -0.3730856264538422 & -0.33763156098430686 & -0.4843753712139662 & -0.5468853635692066\\
$\mathbb{M}_{\mathrm{CP\ }\alpha=0.2}$ & 0.2794935447888979 & 0.24002845554487418 & 0.24319647277616122 & 0.26341961101858224 & 0.06694688046737106\\
\bottomrule
\end{tabular}
}
\caption{1B models, hard labels, datapoints where most models fail}
\label{tab:dynasent balanced:hard:corrtable-models-1b-fail}
    \end{subfigure}\quad
    \begin{subfigure}[t]{0.475\linewidth}
        \centering
        \resizebox{\textwidth}{!}{
\begin{tabular}{l  S[table-format=2.4] S[table-format=2.4]  S[table-format=2.4] S[table-format=2.4]  S[table-format=2.4] } %r rrr r rrrr} % <================================ p{6cm}
\toprule
{} & $\mathbb{M}_\mathrm{fail}^{\mathrm{ref}}$ & $\mathbb{M}_\mathrm{1^{st}\ layer}^{\mathrm{ref}}$ & $\mathbb{M}_\mathrm{1^{st}\ ckpt}^{\mathrm{ref}}$ & $\mathbb{M}_\mathrm{avg\ ckpt}^{\mathrm{ref}}$ & $\mathbb{M}_{\mathrm{avg\ ckpt\ }p}^{\mathrm{ref}}$  \\ 
\midrule
$\mathbb{M}_\mathrm{dis}$ & 0.8521893326955964 & 0.7383351452703243 & 0.8302526961388482 & 0.8230682619738556 & 0.8083794928008469\\
$\mathbb{M}_\mathrm{ent}$ & 0.8509831577345254 & 0.7371743593468848 & 0.8288884148502073 & 0.8220047594091243 & 0.8073493827018579\\
$\mathbb{M}_\mathrm{avg\ ent}$ & 0.7860372052879591 & 0.6693254064869678 & 0.8112738937220045 & 0.7769574179819668 & 0.8518970825968609\\
$\mathbb{M}_{\mathrm{CP\ }\alpha=0.05}$ & 0.7091209772415865 & 0.5811809761635419 & 0.7301496386629457 & 0.7073830603342671 & 0.7697469289932992\\
$\mathbb{M}_{\mathrm{CP\ }\alpha=0.1}$ & 0.7794122548775804 & 0.6588220604667656 & 0.8057164491410197 & 0.7729058197879835 & 0.8224712435461665\\
$\mathbb{M}_{\mathrm{CP\ }\alpha=0.2}$ & -0.5254876567041188 & -0.4793619747900604 & -0.5538131046887752 & -0.5171934230993399 & -0.533163756435052\\
\bottomrule
\end{tabular}
}
\caption{1B models, hard labels, datapoints where most models succeed}
\label{tab:dynasent balanced:hard:corrtable-models-1b-succeed}
    \end{subfigure}

\caption{Spearman correlation on DynaSent re-balanced data between reference-dependent and reference-free indicators of data complexity, broken down by average model success or failure.}
\label{tab:dynasent balanced:corrtable-models-partitioned}
\end{table*}
%\vspace{-2cm}

%% file: Tables/dynasent-original/spearman-vs-h-ent-original.tex
\begin{tabular}{l  S[table-format=2.4] S[table-format=2.4] @{\qquad} S[table-format=2.4] S[table-format=2.4] } %r rrr r rrrr} % <================================ p{6cm}
\toprule
 & \multicolumn{2}{c@{\qquad}}{<1B pool} & \multicolumn{2}{c}{1B pool} \\
 & {{soft}} & {{hard}} & {{soft}} & {{hard}} \\
\midrule
$\mathbb{M}_\mathrm{dis}$ &  0.069460  & 0.069930  &  0.138016 &  0.133244   \\
$\mathbb{M}_\mathrm{ent}$ &  0.071170  & 0.080061  &  0.144865 &  0.136778   \\
$\mathbb{M}_\mathrm{avg\ ent}$ &  0.141544  & 0.130062  &  0.208142 &  0.131972   \\
$\mathbb{M}_{\mathrm{CP\ }\alpha=0.05}$ & 0.127812  & 0.126454  &  0.188075 &  0.153301   \\
$\mathbb{M}_{\mathrm{CP\ }\alpha=0.1}$ & 0.129200  & 0.125054  &  0.195179 &  0.155650   \\
$\mathbb{M}_{\mathrm{CP\ }\alpha=0.2}$ & 0.124749  & 0.108697  &  0.029448 &  0.070866   \\
\midrule
$\mathbb{M}_\mathrm{fail}^{\mathrm{ref}}$ & 0.139867  & 0.142633  &  0.170867 &  0.172466   \\
$\mathbb{M}_\mathrm{1^{st}\ layer}^{\mathrm{ref}}$ & 0.123916  & 0.120525  &  0.201257 &  0.184286   \\
$\mathbb{M}_\mathrm{1^{st}\ ckpt}^{\mathrm{ref}}$ &  0.141114  & 0.143787  &  0.198989 &  0.200628   \\
$\mathbb{M}_\mathrm{avg\ ckpt}^{\mathrm{ref}}$ &  0.134444  & 0.137036  &  0.173982 &  0.174868   \\
$\mathbb{M}_{\mathrm{avg\ ckpt\ }p}^{\mathrm{ref}}$ &  0.162732 & 0.160526  &  0.221395 &  0.188518   \\
\bottomrule
\end{tabular}

%% file: Tables/dynasent-original/corrs-model-when-fail.tex
\begin{table*}[ht!]
\centering
    \begin{subfigure}[t]{0.475\linewidth}
        \centering
        \resizebox{\textwidth}{!}{
\begin{tabular}{l  S[table-format=2.4] S[table-format=2.4]  S[table-format=2.4] S[table-format=2.4]  S[table-format=2.4] } %r rrr r rrrr} % <================================ p{6cm}
\toprule
{} & $\mathbb{M}_\mathrm{fail}^{\mathrm{ref}}$ & $\mathbb{M}_\mathrm{1^{st}\ layer}^{\mathrm{ref}}$ & $\mathbb{M}_\mathrm{1^{st}\ ckpt}^{\mathrm{ref}}$ & $\mathbb{M}_\mathrm{avg\ ckpt}^{\mathrm{ref}}$ & $\mathbb{M}_{\mathrm{avg\ ckpt\ }p}^{\mathrm{ref}}$ \\
\midrule 
$\mathbb{M}_\mathrm{dis}$ & -0.7785696656787301 & -0.7574394623270211 & -0.7875021256326801 & -0.7709132709001326 & -0.7040650313949696\\
$\mathbb{M}_\mathrm{ent}$ & -0.7339301869824422 & -0.7201753610566703 & -0.7402278187165585 & -0.7326611472585954 & -0.6448071038321462\\
$\mathbb{M}_\mathrm{avg\ ent}$ & -0.5347498884540622 & -0.5085556568743512 & -0.5059334295452806 & -0.5459698700592067 & -0.5115701611233574\\
$\mathbb{M}_{\mathrm{CP\ }\alpha=0.05}$ & -0.49135674404086943 & -0.4610056311298333 & -0.46725389249865273 & -0.5049060033810289 & -0.46000779854483675\\
$\mathbb{M}_{\mathrm{CP\ }\alpha=0.1}$ & -0.49634254835186836 & -0.47011064196640656 & -0.4671638446713339 & -0.5089776953960736 & -0.4740800777507762\\
$\mathbb{M}_{\mathrm{CP\ }\alpha=0.2}$ & -0.5432611088460195 & -0.528764949303155 & -0.5146336928692161 & -0.5534608201518799 & -0.5469901005172566\\
\bottomrule
\end{tabular}
}
\caption{<1B models, soft labels, datapoints where most models fail}
\label{tab:dynasent original:soft:corrtable-models-bert-fail}
    \end{subfigure}\quad
    \begin{subfigure}[t]{0.475\linewidth}
        \centering
        \resizebox{\textwidth}{!}{
\begin{tabular}{l  S[table-format=2.4] S[table-format=2.4]  S[table-format=2.4] S[table-format=2.4]  S[table-format=2.4] } %r rrr r rrrr} % <================================ p{6cm}
\toprule
{} & $\mathbb{M}_\mathrm{fail}^{\mathrm{ref}}$ & $\mathbb{M}_\mathrm{1^{st}\ layer}^{\mathrm{ref}}$ & $\mathbb{M}_\mathrm{1^{st}\ ckpt}^{\mathrm{ref}}$ & $\mathbb{M}_\mathrm{avg\ ckpt}^{\mathrm{ref}}$ & $\mathbb{M}_{\mathrm{avg\ ckpt\ }p}^{\mathrm{ref}}$  \\ 
\midrule
$\mathbb{M}_\mathrm{dis}$ & 0.9449410001315872 & 0.9099695500527173 & 0.9416380378090409 & 0.9312395417976435 & 0.8864990096569845\\
$\mathbb{M}_\mathrm{ent}$ & 0.9365510192286866 & 0.9105755193738894 & 0.9342191491156788 & 0.9245881673178926 & 0.895264297306289\\
$\mathbb{M}_\mathrm{avg\ ent}$ & 0.7410835588709049 & 0.7965845454411592 & 0.756571127538538 & 0.7283476462140283 & 0.9543952922517395\\
$\mathbb{M}_{\mathrm{CP\ }\alpha=0.05}$ & 0.711450404279107 & 0.7869511726548879 & 0.7242160867043633 & 0.7040064671199365 & 0.9270452849544341\\
$\mathbb{M}_{\mathrm{CP\ }\alpha=0.1}$ & 0.739031007970219 & 0.7973396956370413 & 0.7538502411354899 & 0.7285561303746217 & 0.9452613612495645\\
$\mathbb{M}_{\mathrm{CP\ }\alpha=0.2}$ & 0.7873055508957284 & 0.818911560353312 & 0.8038244411981236 & 0.7734475208152272 & 0.9493099568750719\\
\bottomrule
\end{tabular}
}
\caption{<1B models, soft labels, datapoints where most models succeed}
\label{tab:dynasent original:soft:corrtable-models-bert-succeed}
    \end{subfigure} 

\vspace{0.1cm}

    \begin{subfigure}[t]{0.475\linewidth}
        \centering
        \resizebox{\textwidth}{!}{
\begin{tabular}{l  S[table-format=2.4] S[table-format=2.4]  S[table-format=2.4] S[table-format=2.4]  S[table-format=2.4] } %r rrr r rrrr} % <================================ p{6cm}
\toprule
{} & $\mathbb{M}_\mathrm{fail}^{\mathrm{ref}}$ & $\mathbb{M}_\mathrm{1^{st}\ layer}^{\mathrm{ref}}$ & $\mathbb{M}_\mathrm{1^{st}\ ckpt}^{\mathrm{ref}}$ & $\mathbb{M}_\mathrm{avg\ ckpt}^{\mathrm{ref}}$ & $\mathbb{M}_{\mathrm{avg\ ckpt\ }p}^{\mathrm{ref}}$ \\
\midrule 
$\mathbb{M}_\mathrm{dis}$ & -0.7996578782009813 & -0.7526115974442616 & -0.7957083198377158 & -0.7884809821153239 & -0.721650857333218\\
$\mathbb{M}_\mathrm{ent}$ & -0.7567504597683716 & -0.7154100887235456 & -0.7513749016270426 & -0.7527854149841084 & -0.6679861590193007\\
$\mathbb{M}_\mathrm{avg\ ent}$ & -0.5807282362036179 & -0.5387925991617519 & -0.5483191809449309 & -0.593790369481737 & -0.5507160690699563\\
$\mathbb{M}_{\mathrm{CP\ }\alpha=0.05}$ & -0.5090103514509462 & -0.4578771714078752 & -0.48051731648180424 & -0.5250801633044418 & -0.47490121572317284\\
$\mathbb{M}_{\mathrm{CP\ }\alpha=0.1}$ & -0.5107094781514837 & -0.47209173093397794 & -0.48259532644852393 & -0.5213027526253328 & -0.49437841789546044\\
$\mathbb{M}_{\mathrm{CP\ }\alpha=0.2}$ & -0.5564991885473153 & -0.5299220060831045 & -0.5243878117573577 & -0.5621788650131778 & -0.5710051150929268\\
\bottomrule
\end{tabular}
}
\caption{<1B models, hard labels, datapoints where most models fail}
\label{tab:dynasent original:hard:corrtable-models-bert-fail}
    \end{subfigure}\quad
    \begin{subfigure}[t]{0.475\linewidth}
        \centering
        \resizebox{\textwidth}{!}{
\begin{tabular}{l  S[table-format=2.4] S[table-format=2.4]  S[table-format=2.4] S[table-format=2.4]  S[table-format=2.4] } %r rrr r rrrr} % <================================ p{6cm}
\toprule
{} & $\mathbb{M}_\mathrm{fail}^{\mathrm{ref}}$ & $\mathbb{M}_\mathrm{1^{st}\ layer}^{\mathrm{ref}}$ & $\mathbb{M}_\mathrm{1^{st}\ ckpt}^{\mathrm{ref}}$ & $\mathbb{M}_\mathrm{avg\ ckpt}^{\mathrm{ref}}$ & $\mathbb{M}_{\mathrm{avg\ ckpt\ }p}^{\mathrm{ref}}$  \\ 
\midrule
$\mathbb{M}_\mathrm{dis}$ & 0.9431841752296455 & 0.8761178791530991 & 0.9376124741059173 & 0.9226521239584872 & 0.8878353115328401\\
$\mathbb{M}_\mathrm{ent}$ & 0.9361131009408283 & 0.8816806372629727 & 0.9321093767831987 & 0.918060020676183 & 0.896318798998128\\
$\mathbb{M}_\mathrm{avg\ ent}$ & 0.7924547591951255 & 0.8089617870307976 & 0.8103893564545479 & 0.7837686243469923 & 0.9690606712951737\\
$\mathbb{M}_{\mathrm{CP\ }\alpha=0.05}$ & 0.7427344496162123 & 0.781682460168947 & 0.7595884309141899 & 0.7381226812713652 & 0.9363081438267253\\
$\mathbb{M}_{\mathrm{CP\ }\alpha=0.1}$ & 0.7608253848190465 & 0.7840157020878631 & 0.7787545967347844 & 0.7509050016963033 & 0.9467315211568488\\
$\mathbb{M}_{\mathrm{CP\ }\alpha=0.2}$ & 0.814595191421903 & 0.8096323074347203 & 0.8327250280464997 & 0.798889359525886 & 0.9506075416327063\\
\bottomrule
\end{tabular}
}
\caption{<1B models, hard labels, datapoints where most models succeed}
\label{tab:dynasent original:hard:corrtable-models-bert-succeed}
    \end{subfigure} 

\vspace{0.1cm}

    \begin{subfigure}[t]{0.475\linewidth}
        \centering
        \resizebox{\textwidth}{!}{
\begin{tabular}{l  S[table-format=2.4] S[table-format=2.4]  S[table-format=2.4] S[table-format=2.4]  S[table-format=2.4] } %r rrr r rrrr} % <================================ p{6cm}
\toprule
{} & $\mathbb{M}_\mathrm{fail}^{\mathrm{ref}}$ & $\mathbb{M}_\mathrm{1^{st}\ layer}^{\mathrm{ref}}$ & $\mathbb{M}_\mathrm{1^{st}\ ckpt}^{\mathrm{ref}}$ & $\mathbb{M}_\mathrm{avg\ ckpt}^{\mathrm{ref}}$ & $\mathbb{M}_{\mathrm{avg\ ckpt\ }p}^{\mathrm{ref}}$ \\
\midrule 
$\mathbb{M}_\mathrm{dis}$ & -0.7581371373748507 & -0.6938126119657477 & -0.7590872325014106 & -0.6972791552136294 & -0.46175147954163254\\
$\mathbb{M}_\mathrm{ent}$ & -0.6730929731840669 & -0.6259103453059905 & -0.6855791438859 & -0.6201578327578491 & -0.3616071410041683\\
$\mathbb{M}_\mathrm{avg\ ent}$ & -0.4073351032586492 & -0.315450676913826 & -0.382919948192304 & -0.40302284020765206 & -0.11796525051470572\\
$\mathbb{M}_{\mathrm{CP\ }\alpha=0.05}$ & -0.3710603186802356 & -0.29244074245458324 & -0.35303763832633156 & -0.36666647906853683 & -0.11343810381598171\\
$\mathbb{M}_{\mathrm{CP\ }\alpha=0.1}$ & -0.579437691788969 & -0.4848594996900842 & -0.469057094585568 & -0.5805997199564036 & -0.5777768291872453\\
$\mathbb{M}_{\mathrm{CP\ }\alpha=0.2}$ & -0.0654292875266202 & -0.08623619588597717 & 0.010342620206129685 & -0.06632250204563839 & -0.3327324397530655\\
\bottomrule
\end{tabular}
}
\caption{1B models, soft labels, datapoints where most models fail}
\label{tab:dynasent original:soft:corrtable-models-1b-fail}
    \end{subfigure}\quad
    \begin{subfigure}[t]{0.475\linewidth}
        \centering
        \resizebox{\textwidth}{!}{
\begin{tabular}{l  S[table-format=2.4] S[table-format=2.4]  S[table-format=2.4] S[table-format=2.4]  S[table-format=2.4] } %r rrr r rrrr} % <================================ p{6cm}
\toprule
{} & $\mathbb{M}_\mathrm{fail}^{\mathrm{ref}}$ & $\mathbb{M}_\mathrm{1^{st}\ layer}^{\mathrm{ref}}$ & $\mathbb{M}_\mathrm{1^{st}\ ckpt}^{\mathrm{ref}}$ & $\mathbb{M}_\mathrm{avg\ ckpt}^{\mathrm{ref}}$ & $\mathbb{M}_{\mathrm{avg\ ckpt\ }p}^{\mathrm{ref}}$  \\ 
\midrule
$\mathbb{M}_\mathrm{dis}$ & 0.8671746547946589 & 0.6965097503305693 & 0.8362402166110658 & 0.8317866347171934 & 0.7919896308012154\\
$\mathbb{M}_\mathrm{ent}$ & 0.8653919115705215 & 0.6955876171800037 & 0.8341858648180448 & 0.8303874261886851 & 0.7923063531760307\\
$\mathbb{M}_\mathrm{avg\ ent}$ & 0.7803025461318012 & 0.7604615167361055 & 0.8098854120393874 & 0.7798943584789388 & 0.9346226406034694\\
$\mathbb{M}_{\mathrm{CP\ }\alpha=0.05}$ & 0.7379881110288817 & 0.700686034675815 & 0.768745021662949 & 0.737746319495458 & 0.8974854550531327\\
$\mathbb{M}_{\mathrm{CP\ }\alpha=0.1}$ & 0.7745996959630642 & 0.7022846903180091 & 0.819862141668393 & 0.7699571172535049 & 0.8822588049782194\\
$\mathbb{M}_{\mathrm{CP\ }\alpha=0.2}$ & 0.1685951369114074 & 0.08564776444488284 & 0.19697128632139344 & 0.16467271427721575 & 0.13961083109103914\\
\bottomrule
\end{tabular}
}
\caption{1B models, soft labels, datapoints where most models succeed}
\label{tab:dynasent original:soft:corrtable-models-1b-succeed}
    \end{subfigure}

\vspace{0.1cm}

    \begin{subfigure}[t]{0.475\linewidth}
        \centering
        \resizebox{\textwidth}{!}{
\begin{tabular}{l  S[table-format=2.4] S[table-format=2.4]  S[table-format=2.4] S[table-format=2.4]  S[table-format=2.4] } %r rrr r rrrr} % <================================ p{6cm}
\toprule
{} & $\mathbb{M}_\mathrm{fail}^{\mathrm{ref}}$ & $\mathbb{M}_\mathrm{1^{st}\ layer}^{\mathrm{ref}}$ & $\mathbb{M}_\mathrm{1^{st}\ ckpt}^{\mathrm{ref}}$ & $\mathbb{M}_\mathrm{avg\ ckpt}^{\mathrm{ref}}$ & $\mathbb{M}_{\mathrm{avg\ ckpt\ }p}^{\mathrm{ref}}$ \\
\midrule 
$\mathbb{M}_\mathrm{dis}$ & -0.7523236199317664 & -0.723247692121742 & -0.7449629305879693 & -0.7108814454897368 & -0.6883336437801221\\
$\mathbb{M}_\mathrm{ent}$ & -0.6753629104338593 & -0.6490483805521672 & -0.6772378319468129 & -0.6429974097439349 & -0.6103760561597104\\
$\mathbb{M}_\mathrm{avg\ ent}$ & -0.4985388181635035 & -0.41976460430631896 & -0.39432049539656044 & -0.4953356749156645 & -0.48143525652000224\\
$\mathbb{M}_{\mathrm{CP\ }\alpha=0.05}$ & -0.42278853339634237 & -0.3493120650833623 & -0.36630251948548287 & -0.4268434605144826 & -0.4001206856173324\\
$\mathbb{M}_{\mathrm{CP\ }\alpha=0.1}$ & -0.4656993674430105 & -0.39249660525790153 & -0.3998869697150864 & -0.471594041400731 & -0.4440362670646855\\
$\mathbb{M}_{\mathrm{CP\ }\alpha=0.2}$ & -0.30616737291783036 & -0.2508037599705166 & -0.24843718175208027 & -0.30393639152561436 & -0.30903458921368265\\
\bottomrule
\end{tabular}
}
\caption{1B models, hard labels, datapoints where most models fail}
\label{tab:dynasent original:hard:corrtable-models-1b-fail}
    \end{subfigure}\quad
    \begin{subfigure}[t]{0.475\linewidth}
        \centering
        \resizebox{\textwidth}{!}{
\begin{tabular}{l  S[table-format=2.4] S[table-format=2.4]  S[table-format=2.4] S[table-format=2.4]  S[table-format=2.4] } %r rrr r rrrr} % <================================ p{6cm}
\toprule
{} & $\mathbb{M}_\mathrm{fail}^{\mathrm{ref}}$ & $\mathbb{M}_\mathrm{1^{st}\ layer}^{\mathrm{ref}}$ & $\mathbb{M}_\mathrm{1^{st}\ ckpt}^{\mathrm{ref}}$ & $\mathbb{M}_\mathrm{avg\ ckpt}^{\mathrm{ref}}$ & $\mathbb{M}_{\mathrm{avg\ ckpt\ }p}^{\mathrm{ref}}$  \\ 
\midrule
$\mathbb{M}_\mathrm{dis}$ & 0.8573673778650786 & 0.7465524404393533 & 0.8239462950749378 & 0.8220722976208995 & 0.8169946126297846\\
$\mathbb{M}_\mathrm{ent}$ & 0.8549732855795165 & 0.7448644731944258 & 0.8211998667015953 & 0.8196077501863221 & 0.8147917343734774\\
$\mathbb{M}_\mathrm{avg\ ent}$ & 0.7638827286484332 & 0.6357798352965525 & 0.7988469841759526 & 0.7505851577867378 & 0.8091003663857522\\
$\mathbb{M}_{\mathrm{CP\ }\alpha=0.05}$ & 0.7022517796897219 & 0.6103214872478411 & 0.7234336490318815 & 0.7025754179379192 & 0.7357447202818431\\
$\mathbb{M}_{\mathrm{CP\ }\alpha=0.1}$ & 0.7482397955925404 & 0.6241037570989001 & 0.7768988795981355 & 0.7438393006868413 & 0.7883730765032823\\
$\mathbb{M}_{\mathrm{CP\ }\alpha=0.2}$ & 0.482983895350373 & 0.41536928587767086 & 0.497665126615869 & 0.47824774139532505 & 0.48981926023997857\\
\bottomrule
\end{tabular}
}
\caption{1B models, hard labels, datapoints where most models succeed}
\label{tab:dynasent original:hard:corrtable-models-1b-succeed}
    \end{subfigure}

\caption{Spearman correlation on DynaSent data (without label re-balancing) between reference-dependent and reference-free indicators of data complexity, broken down by average model success or failure.}
\label{tab:dynasent original:corrtable-models-partitioned}
\end{table*}
%\vspace{-2cm}